%% file: main.tex
\documentclass[twoside,11pt]{article}

\usepackage{blindtext}

\usepackage{wrapfig}
\usepackage{jmlr2e}
\usepackage{amsmath}
\usepackage{booktabs}
\usepackage{multirow}
\usepackage{subcaption}
\usepackage{graphicx}
\usepackage{lastpage}
\firstpageno{1}

\begin{document}

\title{Physics-Informed Super-Resolution of Atmospheric Data}

\author{\name Chang Xu \email chang.xu@epfl.ch\\
       \addr EPFL, Switzerland
       \AND
       \name Gencer Sumbul \email  \\
       \addr EPFL, Switzerland
       \AND
       \name Hugo Porta \email  \\
       \addr EPFL, Switzerland
        \AND
       \name Manon Béchaz \email  \\
       \addr EPFL, Switzerland
        \AND
        \name Sebastian Schemm \email  \\
       \addr University of Cambridge, England
        \AND
        \name Devis Tuia \email  \\
       \addr EPFL, Switzerland}

\editor{My editor}

\maketitle

\begin{abstract}%   <- trailing '%' for backward compatibility of .sty file
% word limitation: 200!
In the context of global warming, extreme events have become more frequent and intense, making their trustworthy detection and forecasting more important than ever. 
Yet, atmospheric observations lack sufficient spatial resolution, motivating atmospheric data downscaling as a way to reconstruct high-resolution data from coarse observations. This task is now being formulated as a super-resolution (SR) problem with machine learning methods featuring high efficiency. Nevertheless, it remains unclear whether the super-resolved atmospheric data still satisfies fundamental physics governing the Earth system, raising concerns about their trustworthiness in climate-related applications. In this work, we address this challenge by constraining SR models to respect hydrostatic primitive equations that represent multivariate atmospheric physics. First, we propose a Physics-Informed Super-Resolution (PISR) method involving multi-scale physics-informed objectives based on primitive equations. PISR favors the SR outputs to respect these equations and therefore naturally encodes inter-variable relationships. In addition, we propose a metric called Normalized Physical Consistency (NPC) derived from said primitive equations to measure the physical consistency of super-resolved data. Experiments on ERA5, CERRA, and COSMO demonstrate that PISR enhances the reconstruction fidelity by improving physical consistency, SR accuracy, and downstream detection of extreme events, as demonstrated by case studies in heatwaves and extreme winds.

\end{abstract}


\newpage

\section{Introduction}
Global warming is changing the Earth system profoundly, with one of the already visible consequences being the increase in both frequency and intensity of extreme events such as
heatwaves, wildfires, and tropical cyclones \citep{fischer2021increasing}. Accurate fine-scale modeling of such complex processes is essential for applications ranging from regional weather forecasting to decision-making in disaster mitigation, agricultural planning, and water resource management. However, due to the limited density of observations and the high computational cost of numerical simulations, operational climate models are often limited to coarse spatial resolutions \citep{era5_2020_quarter,ridal2024cerra}.

Downscaling of atmospheric fields allows for highly resolved variables and enables a more fine-grained analysis of the Earth dynamics and extreme events. The process usually reconstructs high-resolution variable states from the observed coarse inputs. Traditionally, dynamic downscaling methods address this task by solving complex atmospheric physical equations with numerical methods \citep{giorgi2015dynamicalreview}. 
Recently, statistical downscaling has been formulated as a super-resolution (SR) problem using deep learning methods that learn mappings from coarse to fine resolutions. 
Ranging from deterministic models (e.g., DeepSD~\citep{deepsd_2017_kdd}, EDSR \citep{edsr_2017_cvprw}) to generative models (e.g., ClimateDiffuse~\citep{climatediffuse_2024_arxiv}, SRGAN~\citep{SRGAN_2017_CVPR}), neural networks have demonstrated competitive accuracy while offering much higher efficiency than numerical methods by approximating the input-output relationship from data alone. However, outputs remain physics-agnostic and risk being inconsistent with atmospheric laws, both for single-variable fields and inter-variable relationships when super-resolving multiple ones.

The last point is particularly relevant when results of SR are used for extremes detection or forecasting models: despite the visually plausible results, it is crucial to understand whether data-driven SR outputs remain consistent with the physics governing the atmospheric system \citep{kashinath2021pinn-climate-review}. These relations not only govern how each atmospheric variable distributes in space and time, but also how multiple variables relate to each other.
Indeed, without preserving physical relationships, data-driven atmospheric SR is limited to downscaling each variable independently, deteriorating the reliability of reconstructed data since different atmospheric variables are naturally coupled and weather events can involve the joint evolution of multiple variables \citep{zhao2025exebench}.
To investigate the physical consistency in atmospheric SR, we ground this study in a fundamental physics system that describes atmospheric dynamics: the Hydrostatic Primitive Equations (HPEs). Derived from Navier-Stokes equations, HPEs characterize the evolution and balance of multivariate atmospheric processes (\textit{e.g.}, temperature, wind, pressure) through a set of governing equations.
Based on HPEs, we develop a deep learning-based, physics-informed SR method as well as a scheme to evaluate the physical consistency of the super-resolved atmospheric fields. More broadly, we examine whether the gains can extend beyond SR to downstream extreme event detection.

\begin{figure}
    \centering
    \includegraphics[width=0.98\linewidth]{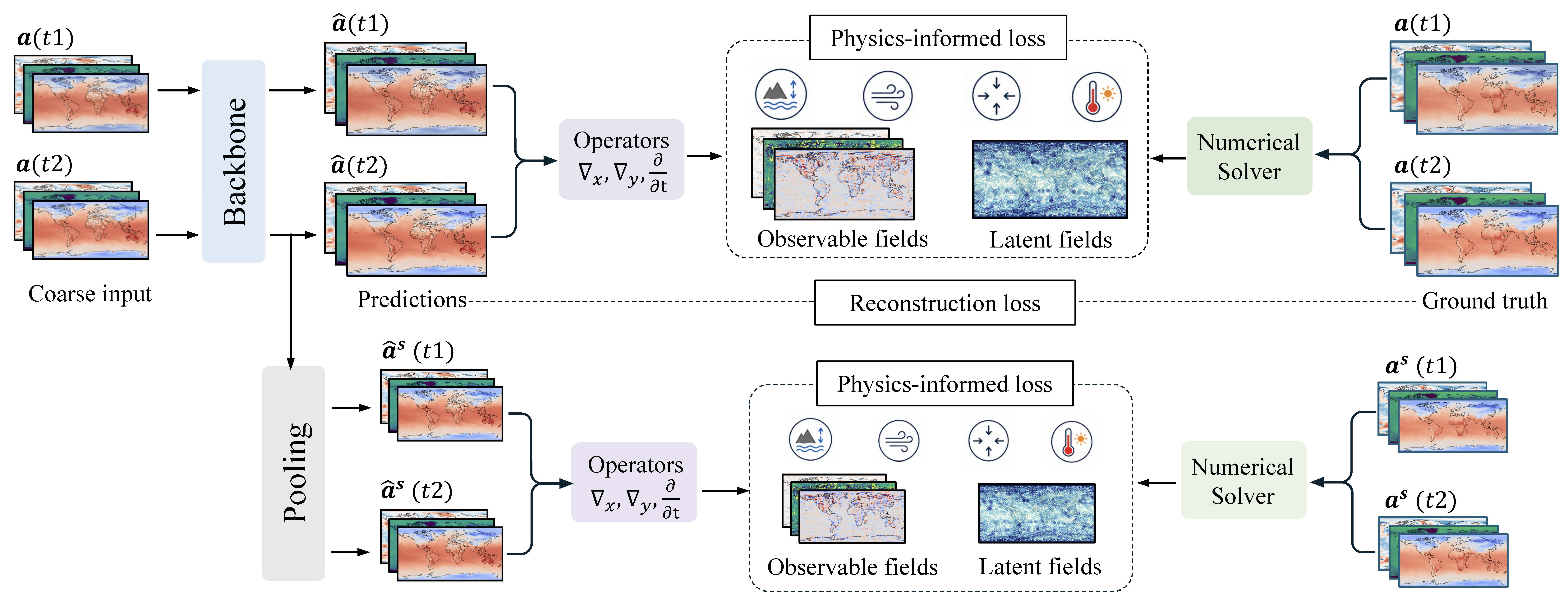}
    \caption{\textbf{Overview of the proposed PISR method.} Given coarse-resolution atmospheric fields at multiple time steps, a backbone network predicts high-resolution atmospheric fields. Differential operators are applied to the prediction to get observable fields $\mathbf{\hat{a}}_o$, while the numerical solver estimates latent fields $\mathbf{a}_l$ from the ground truth. The predictions are supervised by a reconstruction loss and further constrained by physics-informed losses computed from multiple governing equations. The same physics-informed supervision is also applied at a pooled lower-resolution scale, enabling multi-scale physical consistency.}
    \label{fig:pisr}
\end{figure}

The proposed physics-informed super-resolution (PISR) method for atmospheric data is illustrated in Figure~\ref{fig:pisr}. 
Rather than reconstructing each atmospheric variable independently, PISR promotes physically coherent relationships across variables, spatial structures, and temporal dynamics under the physical constraints derived from HPEs.
In PISR, we address two key challenges when constructing the physical constraints: first is the \textit{missing variable problem}: some variables of HPEs, such as friction and diabatic heating, are not directly observed; we therefore treat them as latent physical fields derived from available atmospheric data. Second is the \textit{scale mismatch issue}, since physical relationships may behave differently across spatial scales. We address this issue with a multi-scale physics loss that constrains the reconstructed fields at both the target resolution and the downsampled coarse scales.

To evaluate the physical consistency of the reconstructed high-resolution fields, we go beyond prior works that focus on single physical laws, such as the mass conservation law \citep{hardconstraint_2023_jmlr,verma2024climode} or the kinetic energy of wind \citep{saccardi2025assessing_downscaling}, by proposing a set of HPE-derived Normalized Physical Consistency (NPC) metrics aware of inter-variable relationships. Our metrics quantitatively assess the physical consistency of reconstructed atmospheric fields from the perspectives of hydrostatic relations, horizontal momentum, mass continuity, and thermodynamic equations.

Empirically, we demonstrate that the benefits of informing the SR process with physics extend beyond improved physical consistency and SR accuracy to better extreme event detection. 
In both deterministic and generative baselines, experimental results show that PISR clearly provides better NPC scores under different resolutions (ERA5 at 2.8125$^{\circ}$, CERRA at 11km, and COSMO at 2.2km) with an improvement in reconstructing most atmospheric variables without any additional cost during inference. Importantly, the improvements are transferable to downstream applications, specifically on heatwave and extreme wind detection, implying a promising solution to predicting extreme climate events with physics-informed neural networks.

\section{Related Work}
\subsection{Super-resolution for Atmospheric Downscaling}

Traditional dynamical downscaling methods \citep{tapiador2020regional,downscaling_survey_2024_isprs} use numerical models to resolve regional atmospheric states from global simulations. Although physically interpretable, these methods are computationally expensive.
To reduce the computational cost, data-driven methods formulate atmospheric downscaling as an SR task that learns a mapping from coarse-resolution inputs to fine-resolution targets. DeepSD \citep{deepsd_2017_kdd} is among the first works to apply deep learning to atmospheric downscaling using SRCNN \citep{srcnn_2016_eccv}. Since then, a wide range of methods have been proposed to improve reconstruction quality from different perspectives \citep{downscaling_survey_2024_isprs}. One line of work focuses on improving the representation learning ability of neural networks. For example, GeoFAR \citep{xu2026geofar} addresses the frequency learning bias by introducing geography-informed and frequency-aware representations.
Another direction incorporates domain knowledge about weather and climate into the downscaling process. For example, \cite{physical_downscaling_2024_james} takes the effects of albedo and elevation into consideration for Antarctic surface melt reconstruction, while DeepUrbanDownscale~\citep{deepurbandownscale_2022_jag} leverages urban morphology for urban temperature downscaling. SmCL \citep{hardconstraint_2023_jmlr} introduces hard constraints such as mass conservation for more general climate downscaling tasks.
To model the uncertainty in the SR process, generative methods like GAN \citep{gan_2020_acm} and diffusion \citep{ddpm_2020_nips} have also been adapted to reconstruct high-resolution targets by conditioning the generative process based on coarse climate data \citep{climatediffuse_2024_arxiv,stvd_2024_nips,springenberg2026diffscale}.

Crucially, existing methods tend to downscale each atmospheric variable independently with limited modeling of their interaction. This is detrimental, since atmospheric variables are naturally coupled through physical relationships. A joint modeling of variables is thus important to improve the consistency of the resulting data with real-world physics.
 
\subsection{Physics-aware Neural Networks in Climate}

Recently, there has been growing interest in combining physics with machine learning models for Earth system modeling.
One of the most important directions is physics regularization, in which known physical relationships are incorporated as soft or hard regularization constraints to guide neural networks toward physically consistent solutions. For example, advection and conservation equations have been incorporated as physical constraints in deep learning models to ensure that forecasts remain consistent with the transport and conservation of quantities over time \citep{verma2024climode,sha2025prediction-mass}. Similarly, the diffusion-advection equation is used for air pollution forecasting \citep{liang2025air}, while PS2F-Net \citep{wang2026physics-precipitation} establishes a relationship between moisture and precipitation for precipitation forecasting. 
More related to the atmospheric data downscaling, SmCL \citep{hardconstraint_2023_jmlr} introduces a hard constraint downscaling method that aims to guarantee the mass of super-resolved data to be consistent with the input data, and PhyDL-NWP\citep{phydl-npw-kdd2025} proposes a relationship between arbitrary PDE terms derived from the climate fields to constrain both forecasting and downscaling. In parallel to physics-based regularization, hybrid approaches coupling numerical solvers with neural networks have emerged, combining the physical fidelity of numerical models with the efficiency of data-driven methods.
NeuralGCM \citep{kochkov2024neuralgcm}, for instance, combines a differentiable numerical solver for atmospheric dynamics with a neural network to generate forecasts of deterministic weather, while WeatherGFT \citep{weathergft_2024_nips} generalizes the weather forecasting model to small time scales with physics-AI hybrid modeling. 

Despite the advances of incorporating physics into ML models, most existing work focuses on forecasting, and the integration of physical knowledge into atmospheric downscaling remains limited. The few existing physics-aware downscaling methods often focus on single physical relationships (e.g., mass conservation), whereas real-world atmospheric dynamics are governed by a complex network of interacting constraints (e.g., those described by HPEs). The extent to which current SR methods conform to physical consistency, and whether these multiple physical relationships can enhance the performance of SR, remain unsolved questions.

\section{Methodology}

In this section, we first formulate atmospheric data downscaling as a multivariate SR problem (Section \ref{sec:problem}), then describe the Hydrostatic Primitive Equations used to model the inter-variable physical relationships (Section \ref{sec:hpe}), and finally present our physics-informed super-resolution (PISR) method (Section \ref{sec:pisr}). 

\subsection{Problem Formulation}
\label{sec:problem}
We present the atmospheric data as a multivariate spatiotemporal field as follows:
\begin{equation}
    \mathbf{a}(\lambda,\phi,t)=[a_1(\lambda,\phi,t),\ldots,a_N(\lambda,\phi,t)]
\end{equation}
which is defined over latitude $\lambda\in[-\pi/2,\pi/2]$, longitude $\phi\in[0,2\pi)$, and time $t\in\mathbb{R}$. Each $a_i$ denotes an atmospheric variable such as temperature, wind, humidity, or pressure. Given the traditional low-resolution discretization of observed fields, 
$\mathbf{a}^{\mathrm{LR}}
\in
\mathbb{R}^{T\times N\times H_{\mathrm{LR}}\times W_{\mathrm{LR}}}$,
atmospheric SR aims to recover the corresponding high-resolution state 
$
\mathbf{a}^{\mathrm{HR}}
\in
\mathbb{R}^{T\times N\times H_{\mathrm{HR}}\times W_{\mathrm{HR}}}$, where $H_{\mathrm{HR}}>H_{\mathrm{LR}}$, and $W_{\mathrm{HR}}>W_{\mathrm{LR}}$.

Conventional SR methods learn an independent mapping for each variable with a neural network $\mathcal{F}_{i,\theta}$:
\begin{equation}
    \widehat{a}^{\mathrm{HR}}_i
=\mathcal{F}_{i,\theta}
(a_i^{\mathrm{LR}}) \quad \forall i \in \{1,\ldots,N\}
\end{equation}
The estimator for each variable is obtained by minimizing the expected discrepancy between the predicted and reference high-resolution data:
\begin{equation}
    \mathcal{F}^{\star}_{i,\theta}\in\arg\min_{\mathcal{F}_{i,\theta}}\;
\mathbb{E}_{(a_i^{\mathrm{LR}},a_i^{\mathrm{HR}})\sim\mathcal{D}_i}
[d(
\mathcal{F}_{i,\theta}(a_i^{\mathrm{LR}}),
a_i^{\mathrm{HR}})],
\end{equation}
where $\mathcal{D}_i$ is the corresponding data distribution, and $d(\cdot,\cdot)$ is the loss that measures the discrepancy between the reference and reconstructed high-resolution data on the same spatial grid.

Although this formulation improves the spatial fidelity of individual variables, it does not explicitly model the physical dependencies between variables. However, atmospheric variables are dynamically coupled via physical processes. We therefore define physics-informed atmospheric SR as a multivariate SR problem. Instead of learning independent mappings for each variable, the model reconstructs the full high-resolution atmospheric state jointly as follows:
\begin{equation}
\widehat{\mathbf{a}}^{\mathrm{HR}}=\mathcal{F}_{\theta}(\mathbf{a}^{\mathrm{LR}})=\mathcal{F}_{\theta}
(a^{\mathrm{LR}}_1,\ldots,a^{\mathrm{LR}}_N)
\end{equation}
Ideally, the reconstructed atmospheric state should also satisfy the underlying physical relationships among variables. Let
\begin{equation}
\mathcal{M}_{\mathrm{phys}}=
\{\mathbf{a}: \mathcal{R}_k(\mathbf{a})=0\} \quad \forall k\in\{1,\ldots,K\}
\end{equation}
denote the manifold of atmospheric states satisfying a set of physical constraints, where $\mathcal{R}_k$ is the $k$-th physical relationship among variables. The physics-informed super-resolution task can then be written as
\begin{equation}
\mathcal{F}^{\star}_{\theta}\in\arg\min_{\mathcal{F}_{\theta}}\;
\mathbb{E}_{(\mathbf{a}^{\mathrm{LR}},\mathbf{a}^{\mathrm{HR}})\sim\mathcal{D}}
[d(\mathcal{F}_{\theta}(\mathbf{a}^{\mathrm{LR}}),
\mathbf{a}^{\mathrm{HR}})],
\quad
\mathrm{s.t.}
\quad
\mathcal{F}_{\theta}(\mathbf{a}^{\mathrm{LR}})
\in
\mathcal{M}_{\mathrm{phys}}
\end{equation}

Considering the simplifying assumptions in governing equations and noise in real-world data, we therefore impose the governing equations as soft constraints to encourage the reconstructed fields to approach the physically consistent manifold. In the following section, we will detail the governing equations used in $\mathcal{M}_{\mathrm{phys}}$.

\subsection{Hydrostatic Primitive Equations}
\label{sec:hpe}
\begin{figure}[t]
    \centering
    \begin{subfigure}[t]{0.42\linewidth}
        \centering
        \includegraphics[width=\linewidth]{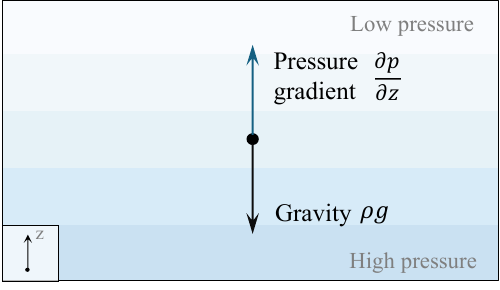}
        \caption{Hydrostatic balance}
        \label{fig:concepts_hydrostatic}
    \end{subfigure}
    \hspace{12pt}
    \begin{subfigure}[t]{0.42\linewidth}
        \centering
        \includegraphics[width=\linewidth]{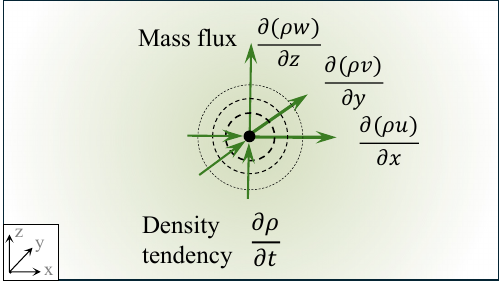}
        \caption{Mass continuity}
        \label{fig:concepts_continuity}
    \end{subfigure}\\
    \vspace{8pt}
    \begin{subfigure}[t]{0.42\linewidth}
        \centering
        \includegraphics[width=\linewidth]{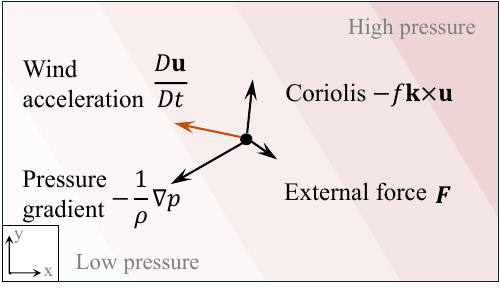}
        \caption{Horizontal momentum}
        \label{fig:concepts_momentum}
    \end{subfigure}
    \hspace{12pt}
    \begin{subfigure}[t]{0.42\linewidth}
        \centering
        \includegraphics[width=\linewidth]{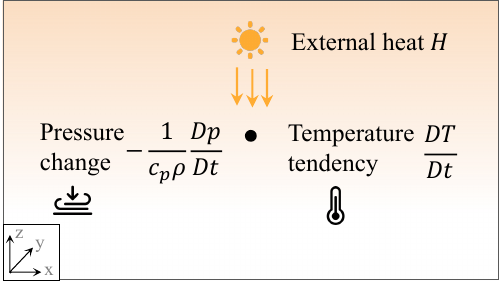}
        \caption{Thermodynamic equation}
        \label{fig:concepts_thermodynamic}
    \end{subfigure}
    \caption{\textbf{Schematic illustration of the hydrostatic primitive equations on a unit parcel.}
    (a) \textit{hydrostatic balance} between the vertical pressure gradient and gravity: $\frac{\partial p}{\partial z}=-\rho g$; 
    (b) \textit{mass continuity} relating the density tendency to the divergence mass fluxes: $\frac{\partial \rho}{\partial t}+\frac{\partial (\rho u)}{\partial x}+\frac{\partial (\rho v)}{\partial y}+\frac{\partial (\rho w)}{\partial z}=0$; 
    (c) \textit{horizontal momentum balance} among wind acceleration, the pressure gradient, the Coriolis force, and external forcing: $\frac{D\mathbf{u}}{Dt}=-f\,\mathbf{k}\times\mathbf{u}-\frac{1}{\rho}\nabla p+\mathbf{F}$; 
    (d) \textit{thermodynamic equation} linking temperature tendency, pressure change, and external diabatic heating: $\frac{D T}{D t}-\frac{1}{c_p\rho}\frac{D p}{D t}=H$.
    Arrows indicate representative directions of the corresponding physical terms acting on the parcel.}
    \label{fig:concepts}
\end{figure}

In this work, the physical constraints are derived from HPEs \citep{charney1955hpe}. HPEs describe atmospheric dynamics and form the basis of many numerical weather and climate models. Figure~\ref{fig:concepts} provides a schematic illustration of the main physical processes represented by HPEs. We first provide a summary of the variables involved and then explain each of the physical processes.
%Therefore, the vertical acceleration is much smaller than the vertical balance be-
%between pressure and gravity, so the hydrostatic approximation is physically reasonable. 

\paragraph{General notions.}
In the below equations, $u$ and $v$ denote the horizontal wind components, $w$ denotes the vertical wind velocity, $p$ denotes the pressure, $\rho$ denotes the air density, $T$ denotes the temperature, $T_v$ denotes the virtual temperature (see Appendix \ref{appendix:tv2m} for its derivation process), $f$ denotes the Coriolis parameter, $g$ denotes the gravitational acceleration, $R$ denotes the gas constant for dry air, $c_p$ denotes the specific heat capacity at constant pressure, $F^{x}$ and $F^{y}$ denote frictions of longitude and latitude directions, $H$ denotes the diabatic heating, and $\frac{D}{Dt}$ denotes the material derivative.
The system of HPEs that we use consists of the following coupled physical relationships.

\paragraph{Hydrostatic equation.}
The hydrostatic equation describes the balance between the vertical pressure-gradient force and gravity as follows:
\begin{equation}
\label{eq:hydro}
\frac{\partial p}{\partial z}
=
-\rho g.
\end{equation}
This equation links the vertical structure of pressure ($\frac{\partial p}{\partial z}$) to air density ($\rho$) under the assumption that vertical acceleration is negligible (\textit{i.e.}, the hydrostatic approximation).
For surface-level variables, we derive a hydrostatic constraint from the hypsometric equation by introducing the unknown sea-level reference pressure as an implicit constant, as detailed in Appendix~\ref{appendix:hydro}. Enforcing this equation encourages the reconstructed atmospheric data to satisfy hydrostatic balance, thereby improving the physical consistency under the hydrostatic assumption. While a limitation of this constraint is that the true atmospheric state may depart from exact hydrostatic balance, especially in fine-scale regions with rapidly evolving vertical motions.

\paragraph{Continuity equation.}
The continuity equation enforces mass conservation as follows:
\begin{equation}
\frac{\partial \rho}{\partial t}+\frac{\partial (\rho u)}{\partial x}+\frac{\partial (\rho v)}{\partial y}+\frac{\partial (\rho w)}{\partial z}=0,
\end{equation}
which constrains the temporal change of density 
\(\left(\frac{\partial \rho}{\partial t}\right)\) 
with the divergence of the atmospheric mass flux. This equation states that air mass cannot be created or destroyed: any local increase or decrease in density must be balanced by the convergence or divergence of mass transport in various directions. In our setting, this equation serves as a mass-consistency constraint to regularize the reconstructed wind, pressure, and density-associated temperature fields.

% \paragraph{Horizontal momentum equations.}
% The horizontal momentum equations describe the evolution of the horizontal wind field under advection, Coriolis force, pressure-gradient force, and unresolved external frictions, as follows:
% \begin{align}
% \frac{D u}{D t} - f v&=-\frac{1}{\rho}\frac{\partial p}{\partial x}+ F^x;
% \\
% \frac{D v}{D t} + f u&=-\frac{1}{\rho}\frac{\partial p}{\partial y}+ F^y.
% \end{align}
% These equations constrain whether the reconstructed wind field ($\frac{D u}{D t}$, $\frac{D v}{D t}$) is dynamically compatible with the reconstructed pressure gradient ($\frac{\partial p}{\partial x}$, $\frac{\partial p}{\partial y}$), Earth's rotation ($fu$, $fv$), and external frictions ($F_x, F_y$).
\paragraph{Horizontal momentum equation.}
The horizontal momentum equation describes the evolution of the horizontal wind field under advection, Coriolis force, pressure-gradient force, and unresolved external friction:
\begin{equation}
\frac{D\mathbf{u}}{Dt}=
-f\,\mathbf{k}\times\mathbf{u}
-\frac{1}{\rho}\nabla p
+\mathbf{F},
\end{equation}
where $\mathbf{u}=(u,v)$ is the horizontal wind vector, $\nabla=(\partial_x,\partial_y)$ is the horizontal gradient operator, $\mathbf{k}$ is the vertical unit vector, and $\mathbf{F}=(F^x,F^y)$ denotes external forcings. This equation constrains whether the reconstructed horizontal wind acceleration is dynamically balanced by the reconstructed pressure gradient, Earth's rotation, and external forcings like frictional effects.

% stopped here
\paragraph{Thermodynamic equation.}
The thermodynamic equation describes the evolution of temperature under advection, compression or expansion, and diabatic heating, as follows:
\begin{equation}
\frac{D T}{D t}-\frac{1}{c_p\rho}\frac{D p}{D t}=H.
\end{equation}
It constrains the consistency between temperature tendency ($\frac{D T}{D t}$), pressure tendency ($\frac{D p}{D t}$), and heat forcing $H$. In the context of atmospheric SR, this equation regularizes whether the reconstructed temperature field evolves consistently with the reconstructed pressure fields.

\paragraph{Equation of state.}
Finally, the equation of state connects pressure, density, and virtual temperature through the ideal-gas relationship:
\begin{equation}
p = \rho R T_v.
\end{equation}
This equation provides the thermodynamic closure of the system. In our setting, we don't directly use this equation as a constraint of the model; instead, we use it to estimate  $\rho$ from pressure $p$ and virtual temperature $T_v$, since density is not directly available as an observed variable.

\subsection{Physics-Informed Super-Resolution (PISR)}
\label{sec:pisr}
\paragraph{Physical constraints.}
PISR reconstructs a multivariate high-resolution atmospheric state
\(\widehat{\mathbf{a}}^{\mathrm{HR}}\), involving most of the observable variables required by the HPEs described in Section~\ref{sec:hpe}. 
However, the full system of HPEs cannot be solved directly in real-world SR settings because several variables, including density \(\rho\), external forcing \(F^x,F^y\), diabatic heating \(H\), and vertical velocity \(w\), are not observed directly.

To address this issue, we leverage HPEs as soft physical constraints rather than solving the corresponding differential equations. We separate the variables involved in HPEs into observable variables $\mathbf{\hat{a}}_{o}$ and latent physical variables $\mathbf{a}_{l}$. The observable variables are obtained from the reconstructed high-resolution fields, while the latent variables are numerically resolved from the corresponding ground-truth high-resolution fields. By doing so, the latent variables provide physically meaningful proxies for the missing terms, allowing us to construct HPEs for physics-informed learning without explicitly solving the full atmospheric system.

Specifically, we define a physics residual operator for each physical relationship:
\begin{equation}
\label{eq.r_hpe}
\mathbf{R}_{\mathrm{HPEs}}
(\mathbf{\hat{a}}_{o}, \mathbf{a}_{l})
=
\left[
\mathcal{R}_{\mathrm{hydro}},\,
\mathcal{R}_{\mathrm{mass}},\,
\mathcal{R}_{\mathrm{mom}},\,
\mathcal{R}_{\mathrm{thermo}}
\right],
\end{equation}
where
$\mathcal{R}_{\mathrm{hydro}}$,
$\mathcal{R}_{\mathrm{mass}}$,
$\mathcal{R}_{\mathrm{mom}}$, and
$\mathcal{R}_{\mathrm{thermo}}$
denote the residuals associated with the hydrostatic, mass-continuity, horizontal-momentum, and thermodynamic relationships, respectively.
Specifically, the residual of the $k$-th physical relationship is defined as
\begin{equation}
\label{eq.general_residual}
\mathcal{R}_{k}
=\mathcal{F}_{k}
(\mathbf{\hat{a}}_{o,k},
\mathbf{a}_{l,k}),
\qquad
k \in
\{
\mathrm{hydro},
\mathrm{mass},
\mathrm{mom},
\mathrm{thermo}
\},
\end{equation}
where $\mathcal{F}_{k}$ denotes the corresponding HPE-derived differential operator, which moves all the terms in each equation to one side. The observable variables $\mathbf{\hat{a}}_{o,k}$ are obtained from the reconstructed high-resolution state $\widehat{\mathbf{a}}^{\mathrm{HR}}$, while
$\mathbf{a}_{l,k}$ contains the latent variables required to close the corresponding physical relationship. These latent variables are not explicitly predicted by the SR model and are instead derived from the ground truth atmospheric state $\mathbf{a}$ according to the corresponding governing equation. We provide the details of physical residual $\mathcal{R}_k$, $\mathbf{\hat{a}}_{o,k}$, and $\mathbf{a}_{l,k}$ for each governing equation in Appendix \ref{appendix:details_rk}.

Rather than explicitly solving the system of HPEs, we force the reconstructed fields to satisfy each governing equation by minimizing the magnitude of its residual. Accordingly, the physical consistency loss associated with the $k$-th relationship is defined as
\begin{equation}
\label{eq.single_physics_loss}
\mathcal{L}_{k}
=
\operatorname{RMSE}
\left(
\mathcal{R}_{k},
0
\right)
=
\sqrt{
\frac{1}{|\Omega|}
\sum_{\Omega}
\left|
\mathcal{R}_{k}
\right|^{2}
},
\end{equation}
where $\Omega$ denotes the spatiotemporal domain over which the residual is evaluated.

\paragraph{Multi-scale learning.}
Scale effects could reduce the effectiveness of a SR system based only on high-resolution constraints. 
Some physical terms, such as rotational balance, are more robust at coarser spatial scales than at the finer spatial scales where SR is operating \citep{holton2013introduction}. 

To mitigate this scale mismatch, we impose physical consistency at multiple spatial scales. Let $\mathcal{P}_s(\cdot)$ denote a non-overlapping average-pooling operator with scale factor $s$, where $s\in\mathcal{S}$ and $\mathcal{S}=\{1,2,\ldots\}$. The reconstructed state at scale $s$ is defined as follows:
\begin{equation}
\widehat{\mathbf{a}}^{(s)}
=
\mathcal{P}_s
\left(
\widehat{\mathbf{a}}^{\mathrm{HR}}
\right).
\end{equation}
For each scale, we first derive the corresponding $\mathbf{\hat{a}}^{(s)}_{o}$ and $\mathbf{a}^{(s)}_{l}$. Then, spatial and temporal derivatives are computed on the pooled fields. The physical constraints from Equation \eqref{eq.r_hpe} can be directly applied to the corresponding resolutions to get the scale-dependent residuals
$\mathbf{R}_{\mathrm{HPE}}(
\mathbf{\hat{a}}^{(s)}_{o}, \mathbf{a}^{(s)}_{l})$. Finally, the combination of multi-scale physics-informed losses over the corresponding HPEs is defined as follows:
\begin{equation}
\mathcal{L}_{\mathrm{phys}}
=
\sum_{s\in\mathcal{S}}
\sum_{k=1}^{K}
\omega_k
\mathcal{L}_{k}^{(s)}, 
\end{equation}
where hyperparameter $\omega_k$ balances the contribution of different equations, and $\mathcal{L}_{k}^{(s)}$ is the $k$-th physics-informed loss at scale $s$.

The final PISR training objective combines the SR reconstruction loss and the multi-scale physics-informed losses:
\begin{equation}
\mathcal{L}_{\mathrm{all}}
=
\mathcal{L}_{\mathrm{rec}}
+
\mathcal{L}_{\mathrm{phys}},
\end{equation}
where \(\mathcal{L}_{\mathrm{rec}}\) measures the discrepancy between the reconstructed and reference high-resolution data, where we employ the Mean Squared Error.

\section{Experiments}

\subsection{Data}

We conduct experiments on three multivariate datasets with increasing spatial resolutions: ERA5 Reanalysis~\citep{era5_2020_quarter}, CERRA~\citep{ridal2024cerra}, and COSMO~\citep{cosmo_model}. These datasets allow us to evaluate atmospheric SR from global-scale coarse fields to regional high-resolution fields.
\begin{itemize}
\item ERA5 is a global reanalysis dataset widely used for weather and climate modeling. Following ClimateLearn \citep{climatelearn_2023_nips}, we construct an hourly dataset covering the period  1981--2018, with a target spatial resolution of \(2.8125^\circ\). 
\item CERRA is a European regional reanalysis produced by ECMWF. We construct a multivariate dataset at \(11\,\mathrm{km}\) target resolution, covering 2010--2021 with a 3-hourly temporal resolution. 
\item COSMO is a high-resolution regional dataset over Switzerland. We use the highest available spatial resolution of \(2.2\,\mathrm{km}\), covering 2015--2020 with hourly temporal resolution. 
\end{itemize}
More details about the datasets can be found in Appendix \ref{appendix:dataset_details}.

\subsection{Experimental Setups}

To assess the performance of existing SR models and the benefit of physics-informed learning, we design experiments evaluating the model's performance across different spatial resolutions and validating the resulting super-resolved data on extreme weather event detection.

\paragraph{Evaluation across scales.}
Evaluating across different spatial scales is important because the properties of atmospheric data may change with resolution, and a model that performs well at one scale may not preserve physical consistency at another. We therefore consider three downscaling settings: \textit{Global downscaling}: ERA5 $(5.625^\circ \rightarrow 2.8125^\circ)$; \textit{Regional downscaling}: CERRA $(22\,\mathrm{km} \rightarrow 11\,\mathrm{km})$; \textit{High-resolution regional downscaling}: COSMO $(17.6\,\mathrm{km} \rightarrow 2.2\,\mathrm{km})$. In addition to the closed-set evaluation, and to assess scale generalization abilities, we also evaluate the model on resolutions unseen during training based on CERRA ($44\,\mathrm{km} \rightarrow 11\,\mathrm{km}$).

\paragraph{Evaluation on extremes.}
Extreme events are particularly important for weather and climate applications, yet machine learning models tend to underestimate rare, high-impact events due to their limited presence in training data \citep{zhang2026physics-outperform-ai}. We therefore evaluate whether physics-informed super-resolved atmospheric fields better detect extreme events. Specifically, we consider: \textit{Heatwaves:} we assess whether the super-resolved T2m improves the detection of heatwave events; \textit{Wind extremes:} we assess whether the super-resolved data improve the detection of extreme wind events (\textit{i.e.}, joint modeling of the $u$ and $v$ components).

We compare our physics-informed models with non-parametric methods, deterministic SR methods, and generative SR methods. More specifically, nonparametric methods include nearest and bilinear interpolation, deterministic neural networks include fundamental architectures (ViT \citep{vit}, U-Net \citep{unet}, EDSR \citep{edsr_2017_cvprw}), task-specific methods (DSFNO \citep{dsfno_2024_jmlr}, SmCL \citep{hardconstraint_2023_jmlr}, DeepSD \citep{deepsd_2017_kdd}), and foundation models (ESFM \citep{ozdemir2026earth}). Finally,  we consider ClimateDiffuse \citep{climatediffuse_2024_arxiv} as a generative SR baseline.  we use EDSR as the PISR baseline for performance comparisons and ViT as the baseline for analysis because of its computational efficiency. More implementation details involving training, evaluation, and variables can be found in Appendix \ref{appendix:implementation}. 

\subsection{Evaluation Metrics}
To evaluate the physical consistency and SR performance, we employ two groups of evaluation metrics. For physical consistency evaluation, we introduce a new metric termed \textit{Normalized Physical Consistency} (NPC), defined as
\begin{equation}
    \mathrm{NPC}_k = \frac{|\mathcal{R}^{\mathrm{pred}}_{k} - \mathcal{R}^{\mathrm{gt}}_{k}|}{|\mathcal{R}^{\mathrm{gt}}_{k}| + \epsilon},
\end{equation}
where \(\mathcal{R}^{\mathrm{pred}}_{k}\) and \(\mathcal{R}^{\mathrm{gt}}_{k}\) denote the physical residuals of the \(k\)-th equation computed from the prediction and ground truth, respectively, \(\epsilon\) is a small positive constant introduced for numerical stability. Physically, NPC quantifies the relative discrepancy between the physical residual of the prediction and that of the ground truth. A lower NPC value indicates that the predicted fields reproduce the physical relationships present in the ground truth more accurately.

For SR quality, we follow previous works~\citep{climatelearn_2023_nips,xu2026geofar} and evaluate the per-pixel reconstruction accuracy using Root Mean Square Error (RMSE).

For downstream extreme event detection, we use the mean intersection over union (mIoU) as the evaluation metric. It measures the overlap between the predicted and ground truth event regions while accounting for both event (foreground) and non-event (background) areas. Specifically, mIoU is calculated as the average of the foreground IoU and background IoU.

\subsection{Physical consistency}
\label{results:physical_consistency}

\begin{table}[t]
\centering
\caption{Physical consistency (NPC $\downarrow$) and SR performance (RMSE $\downarrow$) on the ERA5 dataset. PISR is built based on EDSR. The units of T2m, Sp, U10, V10, and Td2m are K, Pa, m s$^{-1}$, m s$^{-1}$, and K, respectively. The best and second-best results are highlighted in bold and underlined, respectively.}
\label{tab:era5}
\resizebox{\textwidth}{!}{
\begin{tabular}{lccccccccc}
\toprule
\multirow{2}{*}{Method}
& \multicolumn{4}{c}{Physical Consistency}
& \multicolumn{5}{c}{Super-resolution Accuracy (RMSE)} \\
\cmidrule(lr){2-5}
\cmidrule(lr){6-10}
& $\mathrm{NPC}_{hydro}$
& $\mathrm{NPC}_{mass}$
& $\mathrm{NPC}_{mom}$
& $\mathrm{NPC}_{thermo}$
& T2m 
& Sp 
& U10
& V10
& Td2m \\
\midrule
Nearest   & 2.016 & 1.197 & 0.020 & 0.777 & 3.116 & 2820.413 & 2.718 & 2.513 & 2.813 \\
Bilinear  & 1.661 & 0.893 & 0.671 & 0.617 & 2.458 &2401.304 & 2.263 & 2.073 & 2.280 \\
\midrule
U-Net            & 0.041 & \underline{0.495} & \underline{0.006} & 0.598 & 0.994 & 96.723 & \underline{0.955} & 0.924 & 0.817 \\
EDSR              & \underline{0.035} & \textbf{0.462} & \textbf{0.001} & \underline{0.532} & \underline{0.900} & 72.473 & \textbf{0.873} & \textbf{0.851} & \textbf{0.734} \\
DeepSD           & 0.082 & 0.664 & 0.009 & 0.972 & 1.544 & 273.547 & 1.389 & 1.282 & 1.168 \\
SmCL             & 1.472 & 0.923 & 0.010 & 0.764 & 2.287 & 2059.290 & 2.137 & 2.022 & 2.011   \\
DSFNO            & 0.329 & 0.860 & 0.039 & 1.873 & 2.741 & 1074.568 & 1.877 & 1.695 & 2.856 \\
ClimateDiffuse   & 0.047 & 0.741 & \textbf{0.001} & 0.686 & 1.290 & \underline{68.026} & 1.488 &1.411 & 1.037 \\
\midrule
PISR            & \textbf{0.028} &\textbf{0.462} & \textbf{0.001} & \textbf{0.521} & \textbf{0.896} & \textbf{59.480} & \textbf{0.873} & \underline{0.853} & \underline{0.736} \\
\bottomrule
\end{tabular}
}
\end{table}

\begin{table}[t]
\centering
\caption{Physical consistency (NPC $\downarrow$) and super-resolution performance (RMSE $\downarrow$) on the CERRA dataset. The units of T2m, Sp, U10, V10, and Rh2m are K, Pa, m s$^{-1}$, m s$^{-1}$, and \%, respectively. The best and second-best results are highlighted in bold and underlined, respectively.}
\label{tab:cerra}
\resizebox{\textwidth}{!}{
\begin{tabular}{lccccccccc}
\toprule
\multirow{2}{*}{Method}
& \multicolumn{4}{c}{Physical Consistency}
& \multicolumn{5}{c}{Super-resolution Accuracy (RMSE)} \\
\cmidrule(lr){2-5}
\cmidrule(lr){6-10}
& $\mathrm{NPC}_{hydro}$
& $\mathrm{NPC}_{mass}$
& $\mathrm{NPC}_{mom}$
& $\mathrm{NPC}_{thermo}$
& T2m 
& Sp 
& U10
& V10
& Rh2m \\
\midrule
Nearest   & 2.340 & 0.934 & 0.260 & 0.734 & 0.582 & 634.944 & 0.447 & 0.461 & 2.367 \\
Bilinear  & 1.645 & 0.639 & 0.601 & 0.435 & 0.459 & 496.454 & 0.339 & 0.351 & 1.867 \\
\midrule
U-Net            & \underline{0.057} & 0.344 & 0.032 & 0.336 & 0.226 & 42.947 & 0.174 & 0.175 & 1.159 \\
EDSR             & 0.036 & \textbf{0.284} & \underline{0.002} & \underline{0.289} & \underline{0.195} & \underline{27.681} & \textbf{0.144} & \textbf{0.145} & \textbf{1.040} \\
DeepSD           & 0.204 & 0.619 & 0.031 & 0.578 & 0.739 & 144.500 & 0.367 & 0.374 & 1.869 \\
SmCL             & 0.790 & 0.723 & 0.353 & 0.709 & 0.460 & 331.144 & 0.355 & 0.361 & 2.070 \\
DSFNO            & 0.780 & 0.674 & 0.298 & 0.636 & 0.427 & 333.006 & 0.338 & 0.345 & 1.916 \\
ClimateDiffuse   & 0.077 & 0.486 & 0.034 & 0.444 & 0.301 & 59.712 & 0.247 & 0.251 & 1.684 \\
\midrule
PISR           & \textbf{0.022} & \underline{0.296} & \textbf{0.000} & \textbf{0.275} & \textbf{0.194} & \textbf{16.203} & \underline{0.151} & \underline{0.152} & \underline{1.076} \\
\bottomrule
\end{tabular}
}
\end{table}

 We present the results of physical consistency on different datasets in Table \ref{tab:era5} (ERA5), Table \ref{tab:cerra} (CERRA), and Table \ref{tab:cosmo} (COSMO), respectively. 
Several observations can be made regarding physical consistency. First, before applying neural networks (\textit{i.e.,} Nearest, Bilinear), the hydrostatic relation ($\mathrm{NPC}_{hydro}$) shows the largest error between the prediction and the ground truth across all datasets, while the mass conservation ($\mathrm{NPC}_{mass}$), horizontal momentum ($\mathrm{NPC}_{mom}$), and thermodynamic equations ($\mathrm{NPC}_{thermo}$) show relatively lower errors. A possible explanation is that the hydrostatic relation is derived under strong assumptions, like ignoring the vertical acceleration, while real atmospheric fields may not perfectly satisfy these assumptions.

Second, the NPC error increases as the target resolution becomes finer (from global scale in Table~\ref{tab:era5} to local, fine-scale in Table~\ref{tab:cosmo}). This trend is pronounced for the hydrostatic relation ($\mathrm{NPC}_{hydro}$), whose error increases exponentially from the coarse global scale (ERA5) to the fine-scale (COSMO) dataset. In contrast, the errors associated with mass conservation, momentum, and thermodynamic relations increase moderately. This is consistent with the hydrostatic approximation, which is more accurate at larger scales. At finer resolutions, topographic variability and non-hydrostatic motions become more important, making hydrostatic balance harder to preserve and maybe even not desirable.

Lastly, the four physical relationships exhibit different levels of learnability. Across all datasets, neural networks reduce the errors of the hydrostatic ($\mathrm{NPC}_{hydro}$) and momentum ($\mathrm{NPC}_{mom}$) relations easily, while improvements in mass conservation ($\mathrm{NPC}_{mass}$) and thermodynamic consistency ($\mathrm{NPC}_{thermo}$) are relatively smaller. For example, on the CERRA dataset (Table \ref{tab:cerra}), $\mathrm{NPC}_{mom}$ is reduced from 0.601 to near 0.000 while $\mathrm{NPC}_{mom}$ is reduced from 0.435 to 0.275 by introducing the neural network to correct the bilinear interpolated results.
A possible explanation is that the hydrostatic and momentum equations mainly represent balance laws, whereas the continuity and thermodynamic equations are prognostic laws governing the evolution of mass and energy. As a result, the latter involves more complex spatiotemporal processes and may be more difficult for the SR model to optimize.

Overall, these results suggest that preserving physical consistency becomes increasingly challenging at finer spatial resolutions due to the dominance of small-scale dynamics, and that prognostic laws are more difficult to optimize than balance laws because of the complicated spatiotemporal interactions.

\begin{table}[h]
\centering
\caption{Physical consistency (NPC $\downarrow$) and super-resolution performance (RMSE $\downarrow$) on the COSMO dataset. On this dataset, we focus on the application to the foundation model: we compare ESFM fine-tuned with the standard MSE loss against ESFM fine-tuned with both the MSE loss and our proposed multi-scale physics-informed loss. The units of T2m, Sp, U10, V10, and Rh are K, Pa, m s$^{-1}$, m s$^{-1}$, and \%, respectively. }
\label{tab:cosmo}
\resizebox{\textwidth}{!}{
\begin{tabular}{lcccccccccc}
\toprule
\multirow{2}{*}{Method} & \multirow{2}{*}{PISR}
& \multicolumn{4}{c}{Physical Consistency}
& \multicolumn{5}{c}{Super-resolution Accuracy (RMSE)} \\
\cmidrule(lr){3-6}
\cmidrule(lr){7-11}
&
& $\mathrm{NPC}_{hydro}$
& $\mathrm{NPC}_{mass}$
& $\mathrm{NPC}_{mom}$
& $\mathrm{NPC}_{thermo}$
& T2m 
& Sp 
& U10
& V10
& Rh \\
\midrule
Nearest  &  & 87.509 & 1.242 & 0.531 & 1.321 & 1.368 & 1973.865 & 1.146 & 1.141 & 0.017 \\
Bilinear &  & 59.180 & 0.961 & 1.129 & \textbf{0.924} & 1.118 & 1599.130 & 0.957 & 0.956 & \textbf{0.007} \\
\midrule
ESFM     &      & 2.211 & 0.806 & 0.060 & 1.033 & 0.690 & 357.106 & 0.737 & 0.737 & 0.033 \\
ESFM     &  $\checkmark$     & \textbf{1.200} & \textbf{0.794} & \textbf{0.029} & \underline{0.944} & \textbf{0.685} & \textbf{323.268} & \textbf{0.734} & \textbf{0.735} & \underline{0.015} \\
\bottomrule
\end{tabular}
}
\end{table}

\subsection{Model comparison} 
\paragraph{Results across scales.}

\begin{wraptable}{r}{0.52\textwidth}
\vspace{-0.3\baselineskip}
\centering
\caption{Scale generalization performance on CERRA (44$\rightarrow$11 km) measured by RMSE ($\downarrow$). The units of T2m, Sp, U10, V10, and Rh2m are K, Pa, m s$^{-1}$, m s$^{-1}$, and \%, respectively.}
\label{tab:cross_scale}
\resizebox{\linewidth}{!}{
\begin{tabular}{lccccc}
\toprule
Method & T2m & Sp & U10 & V10 & Rh2m \\
\midrule
Bilinear & 0.746 & 815.379 & 0.609 & 0.625 & 3.133 \\
ViT      & 0.716 & 515.324 & 0.597 & 0.610 & \textbf{3.062} \\
PISR[ViT]     & \textbf{0.695} & \textbf{490.628} & \textbf{0.596} & \textbf{0.610} & 3.069 \\
\bottomrule
\end{tabular}
}
\vspace{-0.3\baselineskip}
\end{wraptable}

\begin{figure*}[t]
    \centering
    \begin{subfigure}[t]{0.49\textwidth}
        \centering
        \includegraphics[width=\linewidth]{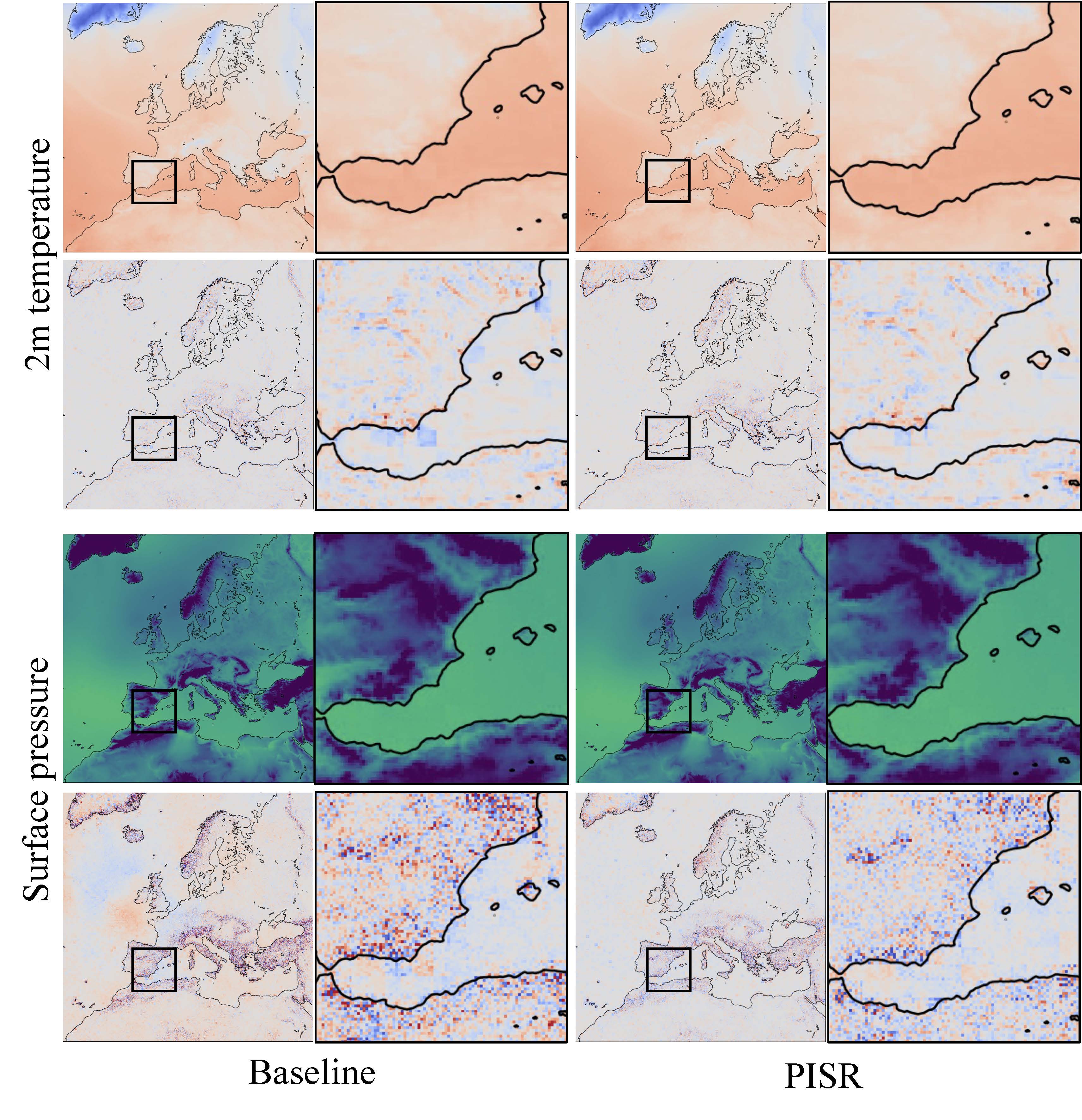}
        \caption{CERRA}
        \label{fig:vis_cerra}
    \end{subfigure}
    \begin{subfigure}[t]{0.5\textwidth}
        \centering
        \includegraphics[width=\linewidth]{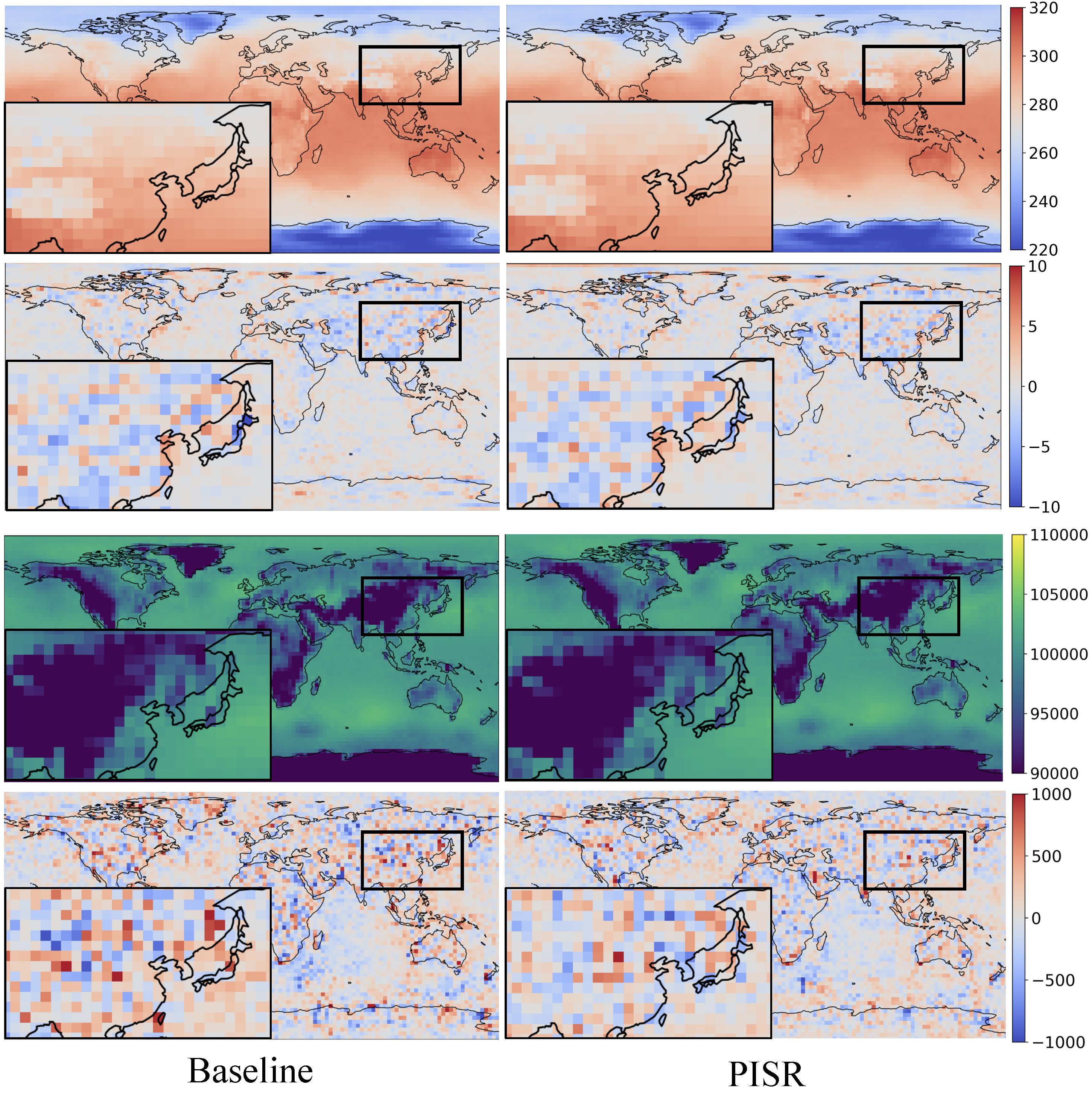}
        \caption{ERA5}
        \label{fig:vis_era5}
    \end{subfigure}
    \caption{
    Qualitative comparison of the super-resolution results produced by the baseline ViT and the PISR on CERRA and ERA5.
    The first and third rows show the reconstructed 2\,m temperature and surface pressure fields, respectively, while the second and fourth rows show the corresponding prediction errors with respect to the ground truth.
    Compared with ViT, PISR better preserves regional spatial structures and produces smaller and more spatially coherent errors, particularly around coastlines and regions with complex terrain.
    }
    \label{fig:sr_visualization}
\end{figure*}

Tables \ref{tab:era5}--\ref{tab:cosmo} compare both the physical consistency and reconstruction ability of different models across datasets. 
From coarse to fine scales, PISR on top of EDSR improves both physical consistency and the SR accuracy of most of the variables. In terms of physics, the largest gains are observed for the hydrostatic ($\mathrm{NPC}_{hydro}$) and momentum ($\mathrm{NPC}_{mom}$) constraints, which are consistent with the observations in Section~\ref{results:physical_consistency}. In terms of reconstruction accuracy, the greatest improvements are achieved for temperature (T2m) and surface pressure (Sp), while the gains for wind components (U10 and V10) and relative humidity (Rh2m) are weaker. 
In Figure \ref{fig:sr_visualization}, we provide a visualization of the SR results on both CERRA and ERA5. Compared to the baseline method, PISR provides a clear improvement in both regional and global datasets by eliminating the prediction bias. Looking into different regions, we observe notable improvements over the ocean, along the coastlines, and in complex terrain, indicating that physics constraint learning can reduce isolated artifacts over the ocean (Figure~\ref{fig:vis_cerra} Bottom), better preserve land–sea transitions (Figure~\ref{fig:vis_cerra}), and thus reconstruct spatial variations more accurately (Figure~\ref{fig:vis_cerra},~\ref{fig:vis_era5}).

Table \ref{tab:cross_scale} explores the generalization potential of PISR across scales, by producing inference on the CERRA dataset, but this time starting from an even coarser resolution of 44km, never seen during training. Results show that, 
even without training on the input resolution, PISR generalizes well to this setup by reducing the errors on T2m, Sp, and U10 by up to 5\% compared to ViT.

\begin{figure}
    \centering
    \includegraphics[width=0.99\linewidth]{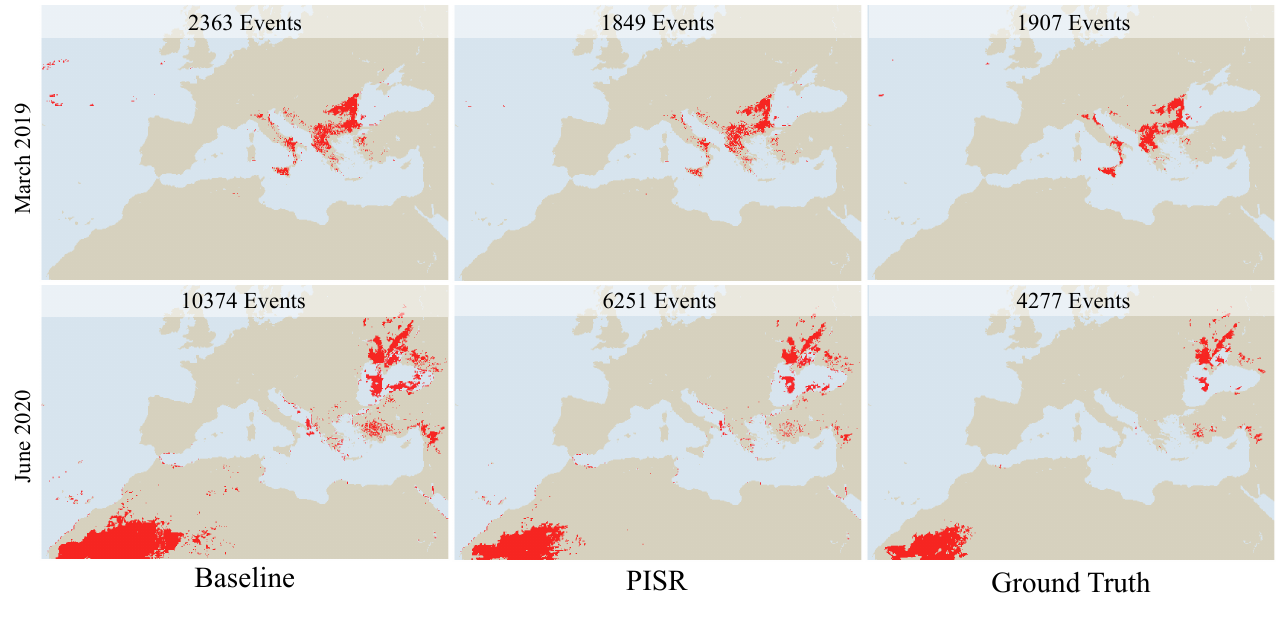}
    \caption{Qualitative comparison of detected heatwaves based on CERRA. Red pixels indicate detected heatwave events. Compared with the baseline ViT model, PISR super-resolved data produces spatial patterns that are visually closer to the ground truth and reduces the number of false detections.}
    \label{fig:heatwave}
\end{figure}

\paragraph{Results on extremes.} Previous studies highlighted that data-driven methods tend to underperform against physics-based numerical methods when forecasting extreme events \citep{zhang2026physics-outperform-ai}. 
We therefore evaluate the utility of super-resolved data in heatwave detection and extreme wind detection. For both events, we compare the IoU obtained when using high-resolution atmospheric data reconstructed by the baseline model and PISR to detect extremes.  
Results in Table \ref{tab:downstream_extremes} show that, compared with the purely data-driven SR, the improvements on extremes are pronounced: the foreground IoU between the detected events and the ground truth is improved by 24\% on heatwaves and 5\% on extreme winds, which is significantly higher than the relative SR improvements of relevant variables (T2m, U10, V10) on CERRA (Table~\ref{tab:architectures}).
Figure~\ref{fig:heatwave} shows heatwave maps derived from the super-resolved atmospheric fields. While both ViT and PISR super-resolved data capture the dominant heatwave region over southeastern Europe, PISR produces spatial patterns that more closely resemble the ground truth and contain fewer isolated detections. These results indicate that improvements in physical consistency can transfer to downstream applications, and physically consistent SR not only improves reconstruction quality but also better preserves the climate signals of extreme events.

\begin{table*}[t]
    \centering
    \caption{
    Evaluation of downstream applications on extreme events. In this table, we compare the detection results using the PISR super-resolved data with respect to its ViT SR baseline. Heatwave and extreme wind detection performance are measured by the mean intersection-over-union (mIoU), foreground IoU (Fg IoU), and background IoU (Bg IoU) between the predicted and ground truth events.
    }
    \label{tab:downstream_extremes}
    \begin{subtable}[t]{0.45\textwidth}
        \centering
        \caption{Heatwave detection}
        \label{tab:heatwave_forecasting}
        \resizebox{\linewidth}{!}{
        \begin{tabular}{lccc}
            \toprule
            SR Data
            & mIoU $\uparrow$
            & Fg IoU $\uparrow$
            & Bg IoU $\uparrow$ \\
            \midrule
            %Bilinear  & -- & -- & -- \\
            ViT       & 74.5 & 50.4 & 98.7 \\
            PISR[ViT]     & \textbf{80.9} & \textbf{62.5} & \textbf{99.2} \\
            \bottomrule
        \end{tabular}
        }
    \end{subtable}
    \hfill
    \begin{subtable}[t]{0.45\textwidth}
        \centering
        \caption{Extreme wind detection}
        \label{tab:heatwave_forecasting}
        \resizebox{\linewidth}{!}{
        \begin{tabular}{lccc}
            \toprule
            SR Data
            & mIoU $\uparrow$
            & Fg IoU $\uparrow$
            & Bg IoU $\uparrow$ \\
            \midrule
            %Bilinear  & -- & -- & -- \\
            ViT       & 86.6 & 73.8 & 99.3 \\
            PISR[ViT]     & \textbf{88.6} & \textbf{77.6} & \textbf{99.5} \\
            \bottomrule
        \end{tabular}
        }
    \end{subtable}
\end{table*}

\subsection{Ablation Study}

\begin{table}[t]
\centering
\caption{Architectural generalization of the proposed method. PISR is compatible with both deterministic models like ViT and EDSR, and generative models like ClimateDiffuse.}
\label{tab:architectures}
\resizebox{\textwidth}{!}{
\begin{tabular}{lccccccccc}
\toprule
\multirow{2}{*}{Method}
& \multicolumn{4}{c}{Physical Consistency}
& \multicolumn{5}{c}{Super-resolution Accuracy (RMSE)} \\
\cmidrule(lr){2-5}
\cmidrule(lr){6-10}
& $\mathrm{NPC}_{hydro}$
& $\mathrm{NPC}_{mass}$
& $\mathrm{NPC}_{mom}$
& $\mathrm{NPC}_{thermo}$
& T2m 
& Sp 
& U10
& V10
& Td2m \\
\midrule
ViT              & 0.335 & 0.676 & 0.023 & 0.618 & 0.506 & 213.755 & 0.359 & 0.360 & 1.938 \\
ClimateDiffuse   & 0.077 & 0.486 & 0.034 & 0.444 & 0.301 & 59.712 & 0.247 & 0.251 & 1.684 \\
EDSR             & 0.036 & 0.284 & 0.002 & 0.289 & 0.195 & 27.681 & 0.144 & 0.145 & 1.040 \\
\midrule
PISR[ViT]              & 0.238 & 0.679 & 0.023 & 0.586 & 0.458 & 175.746 & 0.350 & 0.355 & 1.904 \\
PISR[ClimateDiffuse]   & 0.039 & 0.587 & 0.019 & 0.517 & 0.349 & 25.279 & 0.299 & 0.303 & 1.873 \\
PISR[EDSR]             & 0.022 & 0.296 & 0.000 & 0.275 & 0.194 & 16.203 & 0.151 & 0.152 & 1.076 \\
\bottomrule
\end{tabular}
}
\end{table}

\begin{table}[t]
\centering
\caption{Ablation study of the proposed method, in which the baseline model is a ViT trained with only the super-resolution loss (MSE). Physics denotes the proposed physics-informed loss, and MS denotes the multi-scale loss.}
\label{tab:ablation-main}
\resizebox{\textwidth}{!}{
\begin{tabular}{ccc|cccc|ccccc}
\toprule
\multirow{2}{*}{Baseline} & \multirow{2}{*}{Physics} & \multirow{2}{*}{MS} & \multicolumn{4}{c|}{Physical Consistency} & \multicolumn{5}{c}{SR Accuracy (RMSE)} \\
\cmidrule(lr){4-7} \cmidrule(lr){8-12}
& & & $\mathrm{NPC}_{hydro}$ & $\mathrm{NPC}_{mass}$ & $\mathrm{NPC}_{mom}$ & $\mathrm{NPC}_{thermo}$ & T2m & Sp & U10 & V10 & Rh2m \\
\midrule
$\checkmark$ & $\times$ & $\times$ & 0.335 & \textbf{0.676} & 0.023 & 0.618 & 0.506 & 213.755 & 0.359 & 0.360 & 1.938 \\
$\checkmark$ & $\checkmark$ & $\times$ &   0.297 & 0.678 & 0.007 & 0.590 & 0.491 & 200.380 & 0.357 & 0.360 & 1.934 \\
$\checkmark$ & $\checkmark$ & $\checkmark$  & \textbf{0.238} & 0.679 & \textbf{0.023} & \textbf{0.586} & \textbf{0.458} & \textbf{175.746} & \textbf{0.350} & \textbf{0.355} & \textbf{1.904} \\
\bottomrule
\end{tabular}
}
\end{table}

\paragraph{Scalability to different models.} To assess the method's generalization across different architectures, we build our PISR on three representative architectures: ViT \citep{vit}, EDSR \citep{edsr_2017_cvprw}, and ClimateDiffuse \citep{climatediffuse_2024_arxiv}. These models cover different SR paradigms: ViT and EDSR are trained with deterministic objectives, while ClimateDiffuse is a generative diffusion model. Results in Table \ref{tab:architectures} show that incorporating the proposed physics constraints improves physical consistency across architectures, particularly for the hydrostatic ($\mathrm{NPC}_{hydro}$) and momentum ($\mathrm{NPC}_{mom}$) relations. On deterministic models, PISR improves SR accuracy on all metrics, with the RMSE on surface pressure reduced by almost half, while other variables retain similar performance. The gains are less evident for the generative model.

\paragraph{Effect of individual components.} We perform an analysis on the effectiveness of each proposed strategy on CERRA. Table \ref{tab:ablation-main} delineates the impact of different components: introducing the physics-informed learning significantly reduces the error for both physical consistency and SR. Based on this, performing a multi-scale operation further reduces errors in physical consistency metrics, indicating that some physical relationships may be preserved better at a coarser scale. In addition, we provide a detailed ablation of the effects of each physical constraint in Appendix \ref{appendix:additional_results}.

\section{Conclusion and Outlooks}
Physically consistent and high-fidelity super-resolution (SR) of atmospheric data is crucial for trustworthy applications in extreme events prediction, climate impact assessment, and decision-making. 
In this paper, we introduce a physics-informed SR method for atmospheric data, as well as a new metric to evaluate the physical consistency of  SR results. Experiments on datasets spanning coarse to fine resolutions demonstrate that incorporating the hydrostatic primitive equations as multi-scale constraints into the SR process not only improves the physical consistency of the predicted fields, but also enhances the pixel-level reconstruction fidelity. Notably, the usage of the super-resolved atmospheric data on extreme event detection tasks indicates that the physics-informed learning significantly improves the detectability of heatwaves and extreme wind events.
These findings indicate that combining the complementary benefits of data-driven methods and physical priors helps with the accurate reconstruction of atmospheric data, with great benefits on downstream tasks involving increasingly frequent and intense extreme event prediction, which remains a bottleneck of current data-driven climate models.

A key limitation of this work lies in the governing equations used to constrain the model. The hydrostatic primitive equations rely on simplifying assumptions, such as hydrostatic balance, which becomes less accurate at fine resolutions or in strongly convective settings. Moreover, the constrained variable is limited by the coverage of variables defined in the governing equations. Unconstrained quantities such as precipitation will potentially yield limited improvements. Future work may relax these assumptions according to the characteristics of real-world data, and could develop more general physical representations for variables whose governing equations are incomplete or difficult to formulate explicitly.

\acks{The authors are grateful for the support from CSC, the funding under the Horizon Europe grant 101213369 (DVPS). This work was also supported by a grant from the Swiss National Supercomputing Centre (CSCS) on Alps.}

\include{appendix}

\vskip 0.2in
\bibliography{chang}

\end{document}

%% file: appendix.tex
\newpage

\appendix

\noindent In this appendix, we present the following information:
\begin{itemize}
    \item More details about the HPEs (Section \ref{appendix:theorem}).
    \item Dimensionless analysis of the scales of each term in the HPEs (Section \ref{appendix:scale_analysis}). 
    \item Details of the datasets involved in the paper (Section \ref{appendix:dataset_details}). 
    \item Implementation details (Section \ref{appendix:implementation}).
    \item Ablation study on the contribution of each equation and more visual results (Section \ref{appendix:additional_results}). 
\end{itemize}
\section{More Details on the HPEs}
\label{appendix:theorem}

\subsection{Derivation of the virtual temperature}
\label{appendix:tv2m}

The virtual temperature \(T_v\) is not directly available from the considered data sources. We therefore derive it from the available near-surface variables. 

For ERA5 and COSMO, \(T_v\) is calculated using 2-m air temperature as $T$, surface pressure as $p$, and 2-m dew-point temperature as $T_d$. 
First, the virtual temperature is approximated as
\begin{equation}
T_v=T\left(1+0.61q\right),
\end{equation}
where \(T\) is the 2-m air temperature in Kelvin and \(q\) is the specific humidity.

For ERA5 and COSMO, the actual vapor pressure is derived from the 2-m dew-point temperature:
\begin{equation}
T_{d,c}=T_d-273.15,
\end{equation}
\begin{equation}
e=611.2\exp\left(\frac{17.67T_{d,c}}{T_{d,c}+243.5}\right),
\end{equation}
where \(T_d\) is the 2-m dew-point temperature in Kelvin, \(T_{d,c}\) is the corresponding temperature in degrees Celsius, and \(e\) is the actual vapor pressure in Pa. The specific humidity is then calculated as
\begin{equation}
q=\frac{0.622e}{p_s-0.378e},
\end{equation}
where \(p_s\) is the surface pressure in Pa. Finally, the virtual temperature is obtained as
\begin{equation}
T_v=T\left(1+0.61\frac{0.622e}{p_s-0.378e}\right).
\end{equation}

For CERRA, $T_v$ is calculated using 2-m air temperature as $T$, surface pressure as $p$, and 2-m relative humidity as $RH$. First, the 2-m air temperature is converted to degrees Celsius:
\begin{equation}
T_c=T-273.15.
\end{equation}
The saturation vapor pressure is then calculated as
\begin{equation}
e_s=611.2\exp\left(\frac{17.67T_c}{T_c+243.5}\right).
\end{equation}
Using the relative humidity \(RH\), expressed in percent, the actual vapor pressure is
\begin{equation}
e=\frac{RH}{100}e_s.
\end{equation}
The corresponding mixing ratio is
\begin{equation}
r=\frac{0.622e}{p_s-e},
\end{equation}
and the specific humidity is
\begin{equation}
q=\frac{r}{1+r}.
\end{equation}
Equivalently, the specific humidity can be written directly as
\begin{equation}
q=\frac{0.622e}{p_s-0.378e}.
\end{equation}
The virtual temperature is finally calculated as
\begin{equation}
T_v=T\left(1+0.61q\right).
\end{equation}

\subsection{Implicit sea-level pressure for surface variables}
\label{appendix:hydro}

For atmospheric variables available at multiple vertical levels, hydrostatic consistency can be evaluated directly using the vertical pressure gradient. For surface-level variables, however, observations along the vertical dimension are unavailable. We therefore derive an equivalent hydrostatic constraint by introducing an implicit sea-level reference pressure.

Given the hydrostatic equation
\begin{equation}
\frac{\partial p}{\partial z}=-\rho g,
\end{equation}
we substitute \(\rho=p/(R_dT_v)\) from the ideal gas law into the hydrostatic equation:
\begin{equation}
\frac{\partial p}{\partial z}=-\frac{pg}{R_dT_v},
\end{equation}
where $R_d$ stands for the specific gas constant for dry air. Dividing both sides by $p$ yields
\begin{equation}
\frac{1}{p}\frac{\partial p}{\partial z}=\frac{\partial \ln p}{\partial z}=-\frac{g}{R_dT_v}.
\end{equation}

Integrating between a reference elevation \(z_0\) and the surface elevation \(z_s\) yields the hypsometric relation
\begin{equation}
\ln p_s-\ln p_0
\approx
-\frac{g(z_s-z_0)}{R_d\overline{T}_v},
\end{equation}
where \(p_s\) is the surface pressure, \(p_0\) is the pressure at the reference elevation, and $\overline{T}_v$ is the layer-mean virtual temperature, which is approximated by the 2m virtual temperature $\overline{T}_v$.

Taking sea level as the reference elevation, \(z_0=0\), and rearranging the equation gives
\begin{equation}
\ln p_s
+
\frac{g z_s}
{R_d T_v}
\approx
\ln p_0.
\end{equation}

Because the right-hand side is constant over the spatial domain, a hydrostatically consistent surface field should yield the same inferred sea-level reference pressure at all spatial locations. We therefore train the model by enforcing the spatial variance of the left-hand side to approach zero:
\begin{equation}
\mathcal{R}_{\mathrm{hydro}}
=\operatorname{Var}_{\lambda,\phi}\left[\ln p_s(\lambda,\phi)+\frac{g z_s(\lambda,\phi)}{R_d T_v(\lambda,\phi)}
\right].
\end{equation}
Here, $\mathcal{R}_{\mathrm{hydro}}$ denotes the hydrostatic physics residual (defined in Equation \eqref{eq.general_residual}) used for both model training and evaluation. Here, it is a special case since all variables in this residual equation are observable variables that do not involve latent fields. 
During training, we encourage the $\mathcal{R}_{\mathrm{hydro}}$ to become zero with MSE loss. The reconstructed surface pressure and temperature fields are thereby encouraged to satisfy the hydrostatic relation, without requiring the sea-level reference pressure $p_0$ to be explicitly specified.

\subsection{Details on the physical residual, observable fields, and latent fields}
\label{appendix:details_rk}
In Table \ref{tab:observable_latent_variables} we provide a full list of the physical residual $\mathcal{R}_k$, observable variables $\mathbf{\hat{a}}_o$ that can be predicted, and the latent fields $\mathbf{a}_l$ in each equation. To simplify the computation, we did not explicitly solve the raw latent variable (e.g., $w$) in the equation, we directly computed the corresponding field (e.g., $\frac{\partial w}{\partial z}$) required in the equation.

\begin{table*}[h]
\centering
\caption{Physical residual $\mathcal{R}_k$, observable variables $\mathbf{\hat{a}}_o$ and latent fields $\mathbf{a}_l$ involved in each Hydrostatic Primitive Equation. The density is diagnostically derived as $\rho=p/(R_dT_v)$.}
\label{tab:observable_latent_variables}
\resizebox{\textwidth}{!}{
\begin{tabular}{llll}
\toprule
\textbf{Equation} & \textbf{Physical residual} & \textbf{Observable variables} & \textbf{Latent fields} \\
\midrule
Hydrostatic equation 
& $\displaystyle \mathcal{R}_{\mathrm{hydro}}
= \mathrm{Var}(\mathrm{ln} p+\frac{gz_s}{R_d T_v})$ 
& $p,\;T_v$ 
& $-$ \\
Continuity equation 
& $\displaystyle \mathcal{R}_{\mathrm{mass}}
= \frac{\partial \rho}{\partial t}
+\frac{\partial(\rho u)}{\partial x}
+\frac{\partial(\rho v)}{\partial y}
+\frac{\partial(\rho w)}{\partial z}$ 
& $u,\;v,\;p,\;T_v$ 
& $w$ \\
Horizontal momentum equation 
& $\displaystyle \mathcal{R}_{\mathrm{mom}}
= \frac{D\mathbf{u}}{Dt}
+ f\,\mathbf{k}\times\mathbf{u}
+ \frac{1}{\rho}\nabla p
- \mathbf{F}$ 
& $u,\;v,\;p,\;T_v$ 
& $w,\;\mathbf{F}$ \\
Thermodynamic equation 
& $\displaystyle \mathcal{R}_{\mathrm{thermo}}
= \frac{DT}{Dt}
-\frac{1}{c_p\rho}\frac{Dp}{Dt}
-H$ 
& $u,\;v,\;p,\;T,\;T_v$ 
& $w,\;H$ \\
\bottomrule
\end{tabular}
}
\end{table*}

\section{Dimensionless Analysis of the Physics Constraints}
\label{appendix:scale_analysis}

To better understand the absolute magnitudes of the physical terms used in our learning objective, we perform a dimensionless analysis of each governing relationship. We consider the following nondimensionalized variables:
\begin{equation}
x=L_h\widetilde{x},\qquad y=L_h\widetilde{y},\qquad z=H_z\widetilde{z},\qquad t=\tau\widetilde{t},
\end{equation}
\begin{equation}
u=U\widetilde{u},\qquad v=U\widetilde{v},\qquad w=W\widetilde{w},\qquad \rho=\rho_0\widetilde{\rho},\qquad T=T_0+\Delta T\widetilde{T}.
\end{equation}
 For the ERA5 (2.8125$^{\circ}$) and CERRA (11km) settings used in our experiments, atmospheric structures and patterns extend across hundreds of kilometers at the mesoscale. We therefore adopt a representative horizontal length scale of $L_h=100~\mathrm{km}$. For the vertical direction, we use $H_z=0.1~\mathrm{km}$ as a characteristic scale representative of the lower-atmospheric height intervals in our data. We further take $U=10~\mathrm{m\,s^{-1}}$, $\rho_0=1.2~\mathrm{kg\,m^{-3}}$, $T_0=280~\mathrm{K}$, $\Delta T=10~\mathrm{K}$, $f=10^{-4}~\mathrm{s^{-1}}$, $g=9.81~\mathrm{m\,s^{-2}}$, and $c_p=1004~\mathrm{J\,kg^{-1}\,K^{-1}}$, yielding
where
\begin{equation}
\tau=\frac{L_h}{U}\approx10^{4}~\mathrm{s},
\qquad
W=\frac{UH_z}{L_h}\approx10^{-2}~\mathrm{m\,s^{-1}}.
\end{equation}

\paragraph{Hypsometric relation.}
Combining hydrostatic balance with the ideal-gas equation gives
\begin{equation}
\ln p_s-\ln p_0
\approx
-\frac{g(z_s-z_0)}{R_d\overline{T}_v},
\end{equation}
where $p_{\mathrm{s}}$ is the surface pressure, $p_0$ is the typical sea level pressure, $z_0$ is the reference sea-level elevation (set to 0), and $z_s$ is the true surface elevation. Since $\ln p_0$ is a constant, our implemented residual is:
\begin{equation}
\displaystyle \mathcal{R}_{\mathrm{hydro}}
= \mathrm{Var}(\mathrm{ln} p+\frac{gz_s}{R_d T_v})
\end{equation}
For a representative temperature of \(T_v\approx280~\mathrm{K}\), the term in $\mathrm{Var}()$ varies from approximately 0.01 at 100m to 0.05 at 4km elevations, respectively. The logarithmic pressure-ratio term varies over a comparable range with the opposite sign. The variance thus is of the order of $10^{-4}\sim10^{-3}$.
\paragraph{Continuity equation.}
For the mass-continuity equation
\begin{equation}
\frac{\partial \rho}{\partial t}+\frac{\partial(\rho u)}{\partial x}+\frac{\partial(\rho v)}{\partial y}+\frac{\partial(\rho w)}{\partial z}=0,
\end{equation}
the temporal density tendency has the characteristic scale
\begin{equation}
\frac{\partial\rho}{\partial t}\sim\frac{\rho_0}{\tau}=
\frac{\rho_0U}{L_h}.
\end{equation}
The horizontal mass-flux divergence terms scale as
\begin{equation}
\frac{\partial(\rho u)}{\partial x}
\sim
\frac{\partial(\rho v)}{\partial y}
\sim
\frac{\rho_0U}{L_h},
\end{equation}
while the vertical mass-flux divergence scales as
\begin{equation}
\frac{\partial(\rho w)}{\partial z}
\sim
\frac{\rho_0W}{H_z}.
\end{equation}
Using \(W=UH_z/L_h\), the terms share the characteristic magnitude
\begin{equation}
\frac{\rho_0U}{L_h}
=
\frac{\rho_0W}{H_z}
\approx
1.2\times10^{-4}~\mathrm{kg\,m^{-3}\,s^{-1}}.
\end{equation}

\paragraph{Horizontal momentum equations.}
Consider the zonal momentum equation
\begin{equation}
\frac{Du}{Dt}-fv
=
-\frac{1}{\rho}\frac{\partial p}{\partial x}
+
F^x,
\end{equation}
The local tendency and horizontal advection terms have the characteristic scale
\begin{equation}
\frac{\partial u}{\partial t}
\sim
u\frac{\partial u}{\partial x}
\sim
v\frac{\partial u}{\partial y}
\sim
\frac{U^2}{L_h}
\approx
10^{-3}~\mathrm{m\,s^{-2}}.
\end{equation}
The vertical advection term scales as
\begin{equation}
w\frac{\partial u}{\partial z}
\sim
\frac{WU}{H_z}
=
\frac{U^2}{L_h}
\approx
10^{-3}~\mathrm{m\,s^{-2}}.
\end{equation}
The Coriolis acceleration has the scale
\begin{equation}
fU
\approx
10^{-3}~\mathrm{m\,s^{-2}}.
\end{equation}
Their relative magnitude is characterized by the Rossby number
\begin{equation}
Ro=\frac{U}{fL_h},
\end{equation}
which is approximately one for the selected scales, meaning that inertial and Coriolis effects are comparable. Taking the horizontal pressure perturbation scale as
\begin{equation}
P_h=\rho_0U^2\approx1.2\times10^{2}~\mathrm{Pa},
\end{equation}
the horizontal pressure-gradient acceleration scales as
\begin{equation}
\frac{P_h}{\rho_0L_h}
\sim
\frac{U^2}{L_h}
\approx
10^{-3}~\mathrm{m\,s^{-2}}.
\end{equation}

\paragraph{Thermodynamic equation.}
For the thermodynamic relationship
\begin{equation}
\frac{DT}{Dt}
-
\frac{1}{c_p\rho}\frac{Dp}{Dt}
=
H,
\end{equation}
the local temperature tendency and horizontal temperature-advection terms scale as
\begin{equation}
\frac{\partial T}{\partial t}
\sim
u\frac{\partial T}{\partial x}
\sim
v\frac{\partial T}{\partial y}
\sim
\frac{U\Delta T}{L_h}
\approx
10^{-3}~\mathrm{K\,s^{-1}}.
\end{equation}
The vertical temperature-advection term has the scale
\begin{equation}
w\frac{\partial T}{\partial z}
\sim
\frac{W\Delta T}{H_z}
=
\frac{U\Delta T}{L_h}
\approx
10^{-3}~\mathrm{K\,s^{-1}}.
\end{equation}
For the horizontal pressure perturbation, the pressure-tendency contribution scales as
\begin{equation}
\frac{1}{c_p\rho_0}\frac{UP_h}{L_h}
\approx
10^{-5}~\mathrm{K\,s^{-1}}.
\end{equation}
The contribution associated with the background vertical pressure variation has the characteristic magnitude
\begin{equation}
\frac{gW}{c_p}
\approx
10^{-4}~\mathrm{K\,s^{-1}}.
\end{equation}
These indicate that the representative diabatic-heating scale is
\begin{equation}
H_0=\frac{U\Delta T}{L_h}\approx10^{-3}~\mathrm{K\,s^{-1}}.
\end{equation}

The implemented physical constraints exhibit broadly comparable numerical scales. The hydrostatic consistency loss is typically of order $10^{-4}\sim10^{-3}$, depending on the remaining spatial variability in the hypsometric relation, while the characteristic residual scales are approximately $10^{-4}~\mathrm{kg\,m^{-3}\,s^{-1}}$ for mass continuity, $10^{-3}~\mathrm{m\,s^{-2}}$ for horizontal momentum, and $10^{-3}~\mathrm{K\,s^{-1}}$ for thermodynamics. Although these quantities have different physical units and are not exactly matched in magnitude, they remain within a relatively limited numerical range. We therefore set all physics-loss weights to one, avoiding manually imposed priorities or dataset-specific tuning among the physical relationships. This simple choice was empirically stable across our experiments. Future work may investigate scale-aware or dynamically adaptive weighting strategies based on characteristic residual magnitudes or predictive uncertainty.

\section{Dataset Details}
\label{appendix:dataset_details}

\begin{table}[t]
\centering
\caption{Climate variables available from different datasets used in the paper.}
\label{appendix:variables}
%\small
\setlength{\tabcolsep}{2pt}
\renewcommand{\arraystretch}{1.08}
\begin{minipage}{\textwidth}
\centering
\begin{subtable}[t]{0.32\textwidth}
\centering
\caption{ERA5}
\resizebox{\linewidth}{!}{%
\begin{tabular}{lll}
\toprule
\textbf{Abbr.} & \textbf{Variable} & \textbf{Unit} \\
\midrule
T2m  & 2-m temperature       & K \\
Td2m & 2-m dewpoint          & K \\
Sp   & Surface pressure      & Pa \\
U10  & 10-m eastward wind    & $\mathrm{m\,s^{-1}}$ \\
V10  & 10-m northward wind   & $\mathrm{m\,s^{-1}}$ \\
\bottomrule
\end{tabular}
}
\end{subtable}
\begin{subtable}[t]{0.32\textwidth}
\centering
\caption{CERRA}
\resizebox{\linewidth}{!}{%
\begin{tabular}{lll}
\toprule
\textbf{Abbr.} & \textbf{Variable} & \textbf{Unit} \\
\midrule
T2m  & 2-m temperature       & K \\
Rh2m & 2-m relative humidity & \% \\
Sp   & Surface pressure      & Pa \\
U10  & 10-m eastward wind    & $\mathrm{m\,s^{-1}}$ \\
V10  & 10-m northward wind   & $\mathrm{m\,s^{-1}}$ \\
\bottomrule
\end{tabular}
}
\end{subtable}
\begin{subtable}[t]{0.32\textwidth}
\centering
\caption{COSMO}
\resizebox{\linewidth}{!}{%
\begin{tabular}{lll}
\toprule
\textbf{Abbr.} & \textbf{Variable} & \textbf{Unit} \\
\midrule
T2m & 2-m temperature       & K \\
Rh  & Relative humidity     & \% \\
Sp  & Surface pressure      & Pa \\
U10 & 10-m eastward wind    & $\mathrm{m\,s^{-1}}$ \\
V10 & 10-m northward wind   & $\mathrm{m\,s^{-1}}$ \\
\bottomrule
\end{tabular}
}
\end{subtable}

\end{minipage}%
\end{table}

\noindent\textbf{ERA5 Reanalysis.}
ERA5~\citep{era5_2020_quarter}, developed by the European Centre for Medium-Range Weather Forecasts (ECMWF), reconstructs historical atmospheric and land-surface conditions by assimilating a wide range of observations into the Integrated Forecasting System (IFS). The original product provides hourly global fields from 1979 onward on a $0.25^{\circ}$ grid, including surface variables and atmospheric variables on 37 pressure levels. 
In this paper, we focus on the surface-level variables involved in the hydrostatic primitive equations, including T2m, Td2m, Sp, U10, and V10 (see Table \ref{appendix:variables}).
Following the setup of the ClimateLearn benchmark \citep{climatelearn_2023_nips}, these variables are remapped to global grids at $5.625^{\circ}$ ($32\times64$) and $2.8125^{\circ}$ ($64\times128$) resolution to construct the climate downscaling task. The data retain an hourly temporal resolution. We use 1981--2015 for training, 2016 for validation, and 2017--2018 for testing.

\noindent\textbf{CERRA.}
The Copernicus European Regional Reanalysis (CERRA) is a high-resolution regional reanalysis covering the European domain. Produced by C3S and ECMWF, it has a native horizontal resolution of approximately 5.5 km and a core temporal coverage from September 1984 to June 2021. 
As listed in Table \ref{appendix:variables}, we choose 5 fundamental variables (T2m, Rh2m, Sp, U10, V10) on the surface level involved in the primitive equations in our study. 
Starting from the native $1069\times1069$ grid, we apply bilinear interpolation to construct spatial grids at approximately 11~km ($534\times534$) and 22~km ($267\times267$), thereby enabling training and evaluating super-resolution models. All fields are sampled at 3-hour intervals. Data from 2010--2017 are used for training, while the period 2018--2021 is used for validation and testing.

\noindent\textbf{COSMO.} 
COSMO is a high-resolution regional atmospheric model developed by the Consortium for Small-scale Modeling for high-resolution weather forecasting and research. In this study, we use COSMO data over Switzerland at a native spatial resolution of approximately 2.2 km. As listed in Table \ref{appendix:variables}, we use four surface variables: T2m, Sp, U10, and V10, and one pressure level variable: RH at 963 hPa and 960 hPa. The native data are interpolated to approximately 17.6 km, forming an 8$\times$ super-resolution task from 17.6 km to 2.2 km. All variables are sampled hourly. Data from 2015-12-01 and 2019-12-31 are used for training, and data from 2020-01-01 and 2020-10-28 are used for validation and testing.

\noindent\textbf{Heatwaves.} The heatwave dataset is constructed by using the T2m variable from the CERRA test set covering 2018 to 2021. Specifically, the heatwave is defined as the event during which T2m exceeds the location-specific 95th percentile for at least three consecutive days. Based on this definition, binary heatwave events are generated from both the ground truth and super-resolved temperature data, where heatwave and non-heatwave grid cells are labeled as foreground and background, respectively. 
The super-resolution models trained on CERRA are evaluated directly on this downstream task without additional fine-tuning. Specifically, for each model, we derive heatwave masks from its super-resolved temperature fields and compare them with the ground-truth heatwave masks. Performance is measured using mean intersection over union (mIoU), foreground IoU, and background IoU between the predicted and ground truth heatwave masks.

\noindent\textbf{Extreme Winds.} The extreme wind dataset is also constructed using the U10 and V10 variables from the CERRA test set covering 2018 to 2021. The 10-m wind speed is first computed as $\sqrt{\mathrm{U10}^2+\mathrm{V10}^2}$. An extreme wind event is then defined as a grid cell whose wind speed exceeds the location-specific 98th percentile of the historical wind-speed record. Based on this definition, binary extreme wind masks are generated from both the ground-truth and super-resolved wind fields, where extreme and non-extreme grid cells are labeled as foreground and background, respectively. The super-resolution models trained on CERRA are directly evaluated on this task without additional fine-tuning. Performance is measured using mean intersection over union (mIoU), foreground IoU, and background IoU between the predicted and ground-truth extreme wind masks.

%\noindent\textbf{Tropical Cyclones.} We reuse the dataset provided by ExEBench \citep{zhao2025exebench} to construct the tropical cyclone dataset. We use the 95 tropical cyclone records provided by the ExEBench to retrieve the corresponding fields from ERA5. To align the variable and spatial coverage of the tropical cyclone dataset with our ERA5-trained super-resolution model, we download the same climate variables as our ERA5 setup, and regrid the data to 2.8125$^{\circ}$ and 5.625$^{\circ}$ for downscaling. The resulting dataset is divided into training and test sets using an 8:2 ratio. We then fine-tune the models pretrained on the ERA5 dataset using the cyclone training subset and evaluate their performance on the held-out set on super-resolving Sp, U10, and V10. 

\section{Implementation Details}
\label{appendix:implementation}

\textbf{Training and evaluation.} Experiments on ERA5 and CERRA are conducted on a workstation with one NVIDIA RTX A5500 GPU, while experiments on COSMO are conducted on the SwissAI Supercomputer using 4 NVIDIA GH200 GPUs. For ERA5, we train for 50 epochs with learning rate $2e-4$, weight decay $2e-4$, and batch size 16. For CERRA, we train for 20 epochs using learning rate $2e-4$, weight decay $2e-4$, and a batch size of 2. For COSMO, we fine-tune the ESFM foundation model for 20 epochs using learning rate $2.5e-4$, weight decay $5e-6$, and batch size 8. 
Early stopping is applied in all datasets if the validation loss does not decrease for 5 consecutive epochs.
In PISR, we set the weights of different physical constraints to one by default, we utilize two different scales for the multi-scale loss: one at the target high-resolution, and another one $2\times$ downscaled with loss weights $\omega_k$ all set to one. The $\epsilon$ is set to $1\times10^{-12}$ in NPC.

\textbf{Variable specification.}
We establish the correspondence between the variables in the HPEs and those provided by the datasets as follows: U10 and V10 represent $u$ and $v$, respectively; T2m represents $T$; and Sp represents $p$. We set $g = 9.80065$, $R = 287.05$, and $c_p = 1004.0$. The Coriolis parameter is calculated as $f = 2\omega \sin(\mathrm{lat})$,
where $\omega = 7.292 \times 10^{-5}$ and $\mathrm{lat}$ denotes the latitude in radians. For ERA5, we set $\partial x$, $\partial y$, and $\partial t$ to the target resolutions of $313{,}000~\mathrm{m}$, $313{,}000~\mathrm{m}$, and $3{,}600~\mathrm{s}$, respectively. For CERRA, the corresponding values are $11{,}000~\mathrm{m}$, $11{,}000~\mathrm{m}$, and $10{,}800~\mathrm{s}$, respectively. For COSMO, they are $2{,}200~\mathrm{m}$, $2{,}200~\mathrm{m}$, and $3{,}600~\mathrm{s}$, respectively. All other variables appearing in the HPEs are treated as latent variables and inferred from the known variables.

\section{Additional Experimental Results}
\label{appendix:additional_results}
In this section, we provide 1) the results of a detailed ablation study of the effects of each physical relationship on the final results, and 2) a visualization of the detected extreme wind events. 

Table \ref{tab:ablation-detailed} presents the detailed ablation results of each governing equation based on the CERRA dataset, from which we have the following key observations. 
First, imposing specific governing equations as physics constraints tends to yield the largest improvements in its corresponding NPC metric.  For example, the hydrostatic, horizontal momentum, and thermodynamic constraints reduce $\mathrm{NPC}_{\mathrm{hydro}}$, $\mathrm{NPC}_{\mathrm{mom}}$, and $\mathrm{NPC}_{\mathrm{thermo}}$ from 0.335, 0.023, and 0.618 to 0.248, 0.013, and 0.587, respectively. This agreement indicates that the proposed NPC metrics meaningfully reflect violations of their associated physical relationships. 
Second, enforcing a single physical relationship can adversely affect the consistency of other equations. while jointly constraining all the relationships achieves the best overall performance. For instance, the hydrostatic constraint improves $\mathrm{NPC}_{\mathrm{hydro}}$ but increases both $\mathrm{NPC}_{\mathrm{mass}}$ and $\mathrm{NPC}_{\mathrm{mom}}$. In contrast, jointly imposing all physical constraints with multi-scale regularization provides the most balanced overall performance.
Finally, the thermodynamic constraint alone provides relatively limited benefits: it improves only $\mathrm{NPC}_{\mathrm{thermo}}$ while degrading the consistency of the other equations and increasing the Sp RMSE from 213.755 to 218.684 Pa. One possible explanation is that the thermodynamic equation depends strongly on unresolved diabatic heating and describes a prognostic process with complex spatiotemporal dynamics, which can be difficult to capture from discretely sampled spatial and temporal data.

\begin{table}[t]
\centering
\caption{Ablation study of different physics constraints on CERRA (22km$\rightarrow$11km). Lower values indicate better performance.}
\label{tab:ablation-detailed}
\resizebox{\textwidth}{!}{
\begin{tabular}{ccccc|cccc|ccccc}
\toprule
\multirow{2}{*}{Hydro} & \multirow{2}{*}{Mass} & \multirow{2}{*}{Mom} & \multirow{2}{*}{Thermo} & \multirow{2}{*}{MS} &
\multicolumn{4}{c|}{Physical Consistency} &
\multicolumn{5}{c}{SR Accuracy (RMSE)} \\
\cmidrule(lr){6-9}
\cmidrule(lr){10-14}
& & & & & $\mathrm{NPC}_{hydro}$ & $\mathrm{NPC}_{mass}$ & $\mathrm{NPC}_{mom}$ & $\mathrm{NPC}_{thermo}$ & T2m & Sp & U10 & V10 & Rh2m \\
\midrule
$\times$ & $\times$ & $\times$ & $\times$ & $\times$
& 0.335 & 0.676 & 0.023 & 0.618 & 0.506 & 213.755 & 0.359 & 0.360 & 1.938 \\
$\checkmark$ & $\times$ & $\times$ & $\times$ & $\times$
& 0.248 & 0.681 & 0.052 & 0.611 & 0.462 & 179.068 & 0.352 & 0.353 & 1.906 \\
$\times$ & $\checkmark$ & $\times$ & $\times$ & $\times$
&  0.287 & 0.674 & 0.077 & 0.604 & 0.460 & 188.555 & 0.350 & 0.351 & 1.889 \\
$\times$ & $\times$ & $\checkmark$ & $\times$ & $\times$
& 0.295 & 0.672 & 0.013 & 0.608 & 0.466 & 190.504 & 0.344 & 0.349 & 1.873 \\
$\times$ & $\times$ & $\times$ & $\checkmark$ & $\times$
& 0.345 & 0.679 & 0.059 & 0.587 & 0.487 & 218.684 & 0.356 & 0.360 & 1.889 \\
$\checkmark$ & $\checkmark$ & $\checkmark$ & $\checkmark$ & $\checkmark$
& 0.238 & 0.679 & 0.023 & 0.586 & 0.458 & 175.746 & 0.350 & 0.355 & 1.904 \\

\bottomrule
\end{tabular}
}
\end{table}

\begin{figure}
    \centering
    \includegraphics[width=0.99\linewidth]{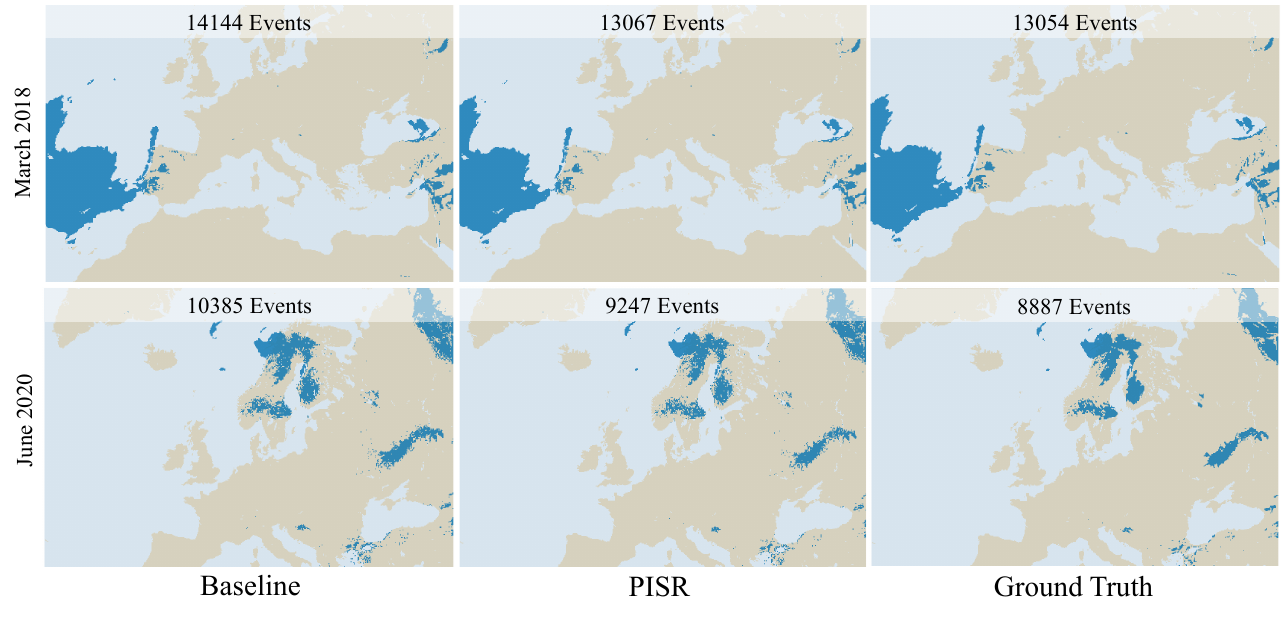}
    \caption{Qualitative comparison of extreme wind detections. Blue pixels indicate detected extreme wind events. Compared with the baseline ViT model, PISR super-resolved data produces spatial patterns that are visually closer to the ground truth and reduce the number of false detections.}
    \label{fig:extreme_wind}
\end{figure}

Figure \ref{fig:extreme_wind} compares the extreme wind detection results from super-resolved data using the baseline ViT model and the ViT-based PISR model. Similar to the heatwave detection results, the data provided by the baseline model tends to lead to an overestimation of the extreme events: in both examples, the ViT baseline yields more than 1000 additional wind extreme events (March 2018: 14144 \textit{v.s.} 13054, June 2020: 10385 \textit{v.s.} 8887) compared with the ground truth results; in contrast, the detections derived from PISR reconstructions are much closer to the ground truth (March 2018: 13067 \textit{v.s.} 13054, June 2020: 9247 \textit{v.s.} 8887). In terms of visual results, the baseline model shows noticeable over-detections on the sea surface (March 2018: west of Ireland, west of Portugal, June 2020: west of Norway), while the PISR more closely approximates the extreme wind patterns derived from the ground truth data.

